%% file: main.tex
\pdfoutput=1

\documentclass[11pt]{article}

\usepackage[preprint]{acl}

\usepackage{times}
\usepackage{latexsym}

\usepackage[T1]{fontenc}

\usepackage[utf8]{inputenc}

\usepackage{microtype}

\usepackage{inconsolata}

\usepackage{graphicx}
\usepackage{amsmath}
\usepackage{amssymb}
\usepackage{bm}
\usepackage{algorithm}
\usepackage{algpseudocode}
\usepackage{url}
\usepackage{booktabs}
\usepackage{siunitx}
\usepackage{multirow}
\usepackage{subcaption}
\usepackage{pifont}
\usepackage{tikz}
\usepackage{tablefootnote}
\usetikzlibrary{positioning, arrows.meta, calc, fit, shadows.blur, decorations.pathreplacing, shapes.callouts}
\usepackage{pgfplots}
\pgfplotsset{compat=1.18}
\usepackage{spverbatim}
\usepackage{tcolorbox}
\tcbuselibrary{listings}

\title{Efficient Context Selection for Long-Context QA: No Tuning, No Iteration, Just Adaptive‑$k$}

\author{Chihiro Taguchi \thanks{Work done during internship at Megagon Labs.}\\
    University of Notre Dame \\
    \texttt{ctaguchi@nd.edu} \\\And
    Seiji Maekawa \\
    Megagon Labs \\
    \texttt{seiji@megagon.ai} \\\And
    Nikita Bhutani \\
    Megagon labs \\
    \texttt{nikita@megagon.ai} \\}

\usepackage{xcolor}

\begin{document}
\maketitle
\begin{abstract}
{Retrieval-augmented generation (RAG) and long-context language models (LCLMs) both address  context limitations of LLMs in open-domain question answering (QA). However, optimal external context to retrieve remains an open problem: fixing the retrieval size risks either wasting tokens or omitting key evidence. Existing adaptive methods like Self-RAG and \textsc{Self-Route} rely on iterative LLM prompting and perform well on factoid QA, but struggle with aggregation QA, where the optimal context size is both unknown and variable.
We present \textbf{Adaptive‑$k$ retrieval}, a simple and effective single-pass method that adaptively selects the number of passages based on the distribution of the similarity scores between the query and the candidate passages.
It does not require model fine-tuning, extra LLM inferences or changes to existing retriever–reader pipelines.
On both factoid and aggregation QA benchmarks, 
Adaptive‑$k$ matches or outperforms fixed‑$k$ baselines while using up to 10× fewer tokens than full-context input, yet still retrieves 70\% of relevant passages.
It improves accuracy across five LCLMs and two embedding models, highlighting that dynamically adjusting context size leads to more efficient and accurate QA.\footnote{The code is available at \url{https://github.com/megagonlabs/adaptive-k-retrieval}.}}

\end{abstract}

\input{sections/01_introduction}
\input{sections/02_related}
\input{sections/03_method}
\input{sections/04_b_experimental}
\input{sections/05_results}
\input{sections/06_conclusion}

\input{sections/07_limitations}
\input{sections/08_ethical}

\section*{Acknowledgments}
We thank Hayate Iso and Pouya Pezeshkpour for their constructive feedback on this study.

\bibliography{custom}

\input{sections/10_appendix}

\end{document}

%% file: sections/01_introduction.tex
\section{Introduction}

Despite remarkable progress in LLMs, efficiently incorporating external knowledge during inference for long or dynamic contexts remains a key challenge.
Two major paradigms have emerged to address this: long-context language models (LCLMs), which extend the model's context window to directly ingest more information, and retrieval-augmented generation (RAG), which retrieves relevant documents from an external corpus to condition the generation.
While these approaches are sometimes presented as alternatives \cite{li2024longcontextvsrag, yu2024defenserageralongcontext}, recent studies highlight their complementary nature \cite{li-etal-2024-retrieval}. 

A central bottleneck in both paradigms is determining how much context to include.
Fixed-size retrieval budgets (\textit{e.g.}, top-$k$ retrieval) are suboptimal, because they either retrieve too little and risk omitting key evidence, or retrieve too much, which can overwhelm the model, increase latency and costs, and degrade performance \cite{yu2024defenserageralongcontext, leng2024longcontextragperformance, jin2024longcontextllmsmeetrag}.
As \citet{yang2024retrievalholisticunderstandingdolce} observes, the challenge in long-context reasoning lies not only in document length but also in how relevant information is distributed and duplicated within the context.
Crucially, query type plays a major role: factoid questions may need only a few targeted facts, while aggregation queries \cite{maekawa2025holisticreasoninglongcontextlms} often require reasoning based on information from multiple evidence spans.
This variability makes fixed-$k$ retrieval suboptimal for complex tasks.

To address this, several hybrid and adaptive retrieval methods such as Self-RAG \cite{asai2023selfraglearningretrievegenerate}, Adaptive-RAG \cite{jeong2024adaptiveraglearningadaptretrievalaugmented}, and Dynamic context cutoff \cite{xie2025knowingstopdynamiccontext} have been proposed, which estimate retrieval depth via iterative prompting, each time fetching a fixed number of documents.
However, they assume white-box access to the LLM: Self-RAG requires fine-tuning the LLM, while dynamic context cutoff depends on access to internal KV cache states. This makes them incompatible with closed-source or API-based LLMs.
While effective on factoid-style questions, they also face significant limitations in terms of scalability, latency, and deployment flexibility.
Although \textsc{Self-Route} \cite{li-etal-2024-retrieval} offers a more modular solution, it still relies on a fixed retrieval size and lacks the ability to adapt to varying information needs across queries and context documents.
This motivates our core research question: \emph{How can we estimate the optimal number of passages to retrieve for a given query and set of context documents, without supervision or iterative prompting?}

\input{src/methods_comparison}

To address this question, we introduce \textbf{Adaptive-$k$ retrieval}, a simple yet effective plug‑and‑play method for dynamically selecting a query- and context-specific number of documents in a single retrieval pass.
Our approach relies on analyzing the distribution of similarity scores between a query and candidate documents.
By identifying the largest gap in the sorted similarity distribution, it estimates an optimal cutoff point, retrieving the top-$k$ documents before the gap.
Unlike prior adaptive retrieval methods, Adaptive‑$k$ requires no model fine-tuning, no access to internal components and no iterative prompting.
It is fully modular, allowing seamless integration with existing retriever–reader pipeline and compatibility with black-box LLMs.
By relying solely on the distributional structure of similarity scores, Adaptive‑$k$ adjusts the retrieval size on a per-query basis. 
This simple yet principled strategy leads to significant reductions in input length and inference cost, while maintaining or even improving the answer quality across both factoid and aggregation-style QA tasks. 
We compare {Adaptive-$k$ retrieval} to prior approaches in Table~\ref{tab:rag}.

We evaluate Adaptive-$k$ on both factoid and aggregation-style QA tasks across multiple LCLMs and embedding models.
Our experiments span two representative long-context benchmarks: HELMET \cite{yen2025helmetevaluatelongcontextlanguage}, which includes factoid QA tasks with up to 128k-token contexts, and HoloBench \cite{maekawa2025holisticreasoninglongcontextlms}, which focuses on aggregation-style queries.
Our results show that on aggregation-QA, Adaptive-$k$ outperforms \textsc{Self-Route} by up to +9 points in answer accuracy on high-information tasks. It consistently maintains $\sim$70\% context recall 
and reduces token usage by 2× to 10× compared to full-context baselines. On factoid QA, Adaptive-$k$ matches or exceeds the accuracy of fixed-size retrieval with up to 99\% reduction in input tokens, effectively pruning irrelevant content. These findings highlight the importance of query-specific context sizing and establish Adaptive-$k$ as a simple, robust, and efficient alternative to more complex adaptive retrieval strategies.

In summary, our key contributions are:

\begin{itemize}
    \item We propose Adaptive-$k$, a simple yet effective plug-and-play method for adaptive document retrieval that dynamically adjusts context size based on similarity distribution statistics. 
    
    {
    \item Adaptive-$k$ achieves higher accuracy than prior methods and up to 99\% token reduction on factoid and aggregation QA against LCLMs with full context.
    
    \item We show that no single fixed-size retrieval strategy fits all settings. In contrast, Adaptive-$k$ shows robust performance across multiple LLMs, embedding models and benchmarks.
    }
\end{itemize}

%% file: src/methods_comparison.tex
\newcommand{\xmark}{\color{red} \ding{55}}
\newcommand{\cmark}{\color{teal} \ding{51}}
\newcommand{\warning}{{\fontencoding{U}\fontfamily{futs}\selectfont\char 49\relax}}

\begin{table*}[t]
    \centering
    \setlength{\tabcolsep}{5pt}
    \newcommand{\modularity}[0]{Plug-and-Play via API}
    \newcommand{\granularity}[0]{Retrieval Amount Variability}
    \newcommand{\cost}[0]{Single Retrieval Operation}
    \scalebox{0.8}{
    \begin{tabular}{lccc} \toprule
           & \modularity & \granularity  & \cost \\ \midrule
        No RAG (LCLM) & \cmark & \xmark & No Retrieval \\
        RAG (traditional) & \cmark & \xmark & \cmark \\
        \midrule
        Self-RAG \cite{asai2023selfraglearningretrievegenerate} & \xmark & \cmark & \xmark \\
        Adaptive-RAG \cite{jeong2024adaptiveraglearningadaptretrievalaugmented} & \cmark & \cmark & \xmark \\
        \textsc{Self-Route} \cite{li-etal-2024-retrieval} & \cmark & \xmark & \cmark \\
        LC-Boost \cite{qian2024longllmsnecessitylongcontexttasks} & \cmark & \cmark & \xmark \\
        Dynamic context cutoff \cite{xie2025knowingstopdynamiccontext} & \xmark & \cmark & \xmark \\
        \midrule
        Adaptive-$k$ RAG (ours) & \cmark & \cmark & \cmark \\
        \bottomrule
    \end{tabular}
    }
    \caption{The comparison of previously proposed approaches as enhanced RAG.
    \emph{\modularity} refers to whether the approach can be easily plugged in to various LLM pipelines.
    \emph{\granularity} refers to whether the system can flexibly change the retrieval amount depending on different queries and context.
    \emph{\cost} refers to whether the retrieval is performed in a single step or in multiple steps.
    }
    \label{tab:rag}
\end{table*}

%% file: sections/02_related.tex
\section{Related Work}

RAG and LCLMs are two prominent paradigms for equipping LLMs with external knowledge. Recent studies show that LCLMs can match or outperform RAG in certain QA tasks \cite{li2024longcontextvsrag, yu2024defenserageralongcontext}, yet the two methods are fundamentally complementary. 

Several approaches have been proposed to leverage both the strengths of RAG and LCLMs with flexible retrieval strategies.
Self-RAG \cite{asai2023selfraglearningretrievegenerate} trains an LLM to generate reflection tokens that enable retrieval on the fly, so that the LLM can determine whether it needs any additional document by itself.
\textsc{Self-Route} \cite{li-etal-2024-retrieval} asks an LLM whether it can answer the query with the retrieved context; if not, the LLM is given the full context.
Adaptive-RAG \cite{jeong2024adaptiveraglearningadaptretrievalaugmented} uses a workflow that iteratively asks an LLM whether it can answer the given query with the retrieved context.
LC-Boost \cite{qian2024longllmsnecessitylongcontexttasks} enables short-context LLMs to tackle long-context tasks by first identifying relevant information, then reasoning over it, without needing extended context windows or fine-tuning. 

While effective in controlled settings, these methods often rely on white-box access to the LLM, fine-tuning, or multiple LLM inferences. Existing research has highlighted key limitations in RAG systems, particularly in terms of cost, modularity, and retrieval granularity. However, prior methods typically address these issues in isolation, and to our knowledge, no single approach has tackled all three challenges holistically. Our method is the first to offer a unified solution that is cost-efficient, modular, and capable of adaptive, query-specific retrieval in a single pass.

\paragraph{Cost.} High-quality inference often comes with high token usage, energy consumption, and latency \cite{li-etal-2024-retrieval, qian2024longllmsnecessitylongcontexttasks}, underscoring the need for more cost-effective alternatives.

\paragraph{Modularity.} 
Modularity is crucial for real-world deployment \cite{wang-etal-2024-searching}, but many existing methods require fine-tuning or training the LLM itself. This tight coupling reduces compatibility with API-based or closed-source models, limiting practical applicability.

\paragraph{Retrieval granularity.}
Aggregation-type queries often require comprehensive evidence and holistic understanding. For example, answering ``Which colleges in California have over 10,000 students?'' demands access to the full set of relevant entries. Fixed-size or iterative retrieval methods struggle with such cases, as they cannot dynamically adjust retrieval depth based on query complexity.

%% file: sections/03_method.tex
\section{Method}

This section details our approach to adaptive retrieval, grounded in the analysis of similarity score patterns to determine retrieval sizes adaptively based on the query and the context. We first review the standard RAG retrieval process, then present our methodology to identify the optimal threshold in similarity distributions to efficiently select relevant documents.

\subsection{Retrieval in vanilla RAG}
RAG consists of two steps: retrieval and generation.
Given a query $q$ and $N$ context documents $C = \{c_i\}_1^N$, the retriever module identifies top-$k$ semantically similar context documents $C'$.
Modern RAG approaches convert the query and the context documents in natural language into the query embedding $\bm{q} \in \mathbb{R}^d$ and context embeddings $\bm{C} \in \mathbb{R}^{N \times d}$. Similarity scores $\bm{s} \in \mathbb{R}^{N}$ are then computed to quantify relevance, commonly using cosine similarity:

$$
\bm{s} = f_\text{sim}(\bm{q}, \bm{C}) = \dfrac{\bm{C}\bm{q}^\top}{||\bm{q}|| \cdot ||\bm{C}||_\text{rows}}
$$

RAG typically retrieves a fixed number of top-$k$ documents (or tokens) based on the practitioner's choice.
This fixed retrieval size is simple and modular but may result in inefficient token usage, either retrieving irrelevant documents or missing critical information, especially when the amount of relevant context varies depending on the provided context documents and the query type.

\begin{figure}[t]
    \centering
    \includegraphics[width=\linewidth]{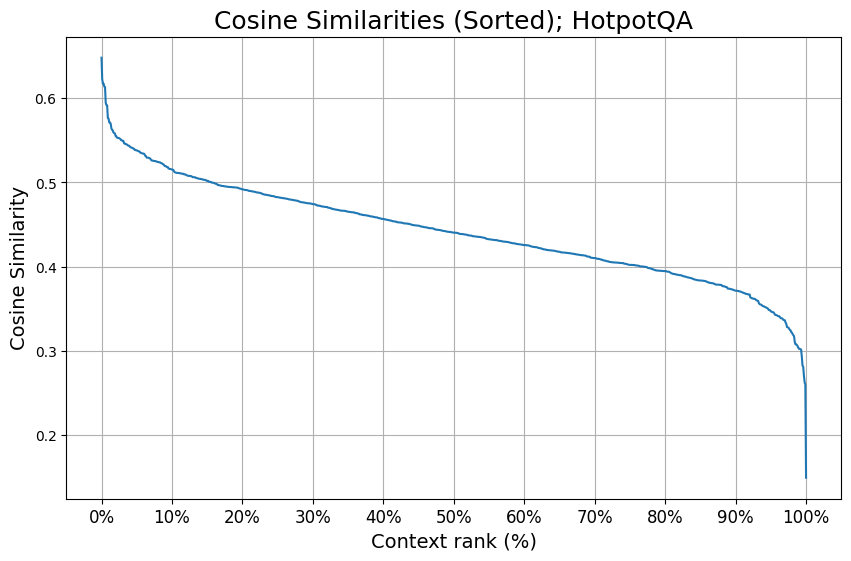}
    \includegraphics[width=\linewidth]{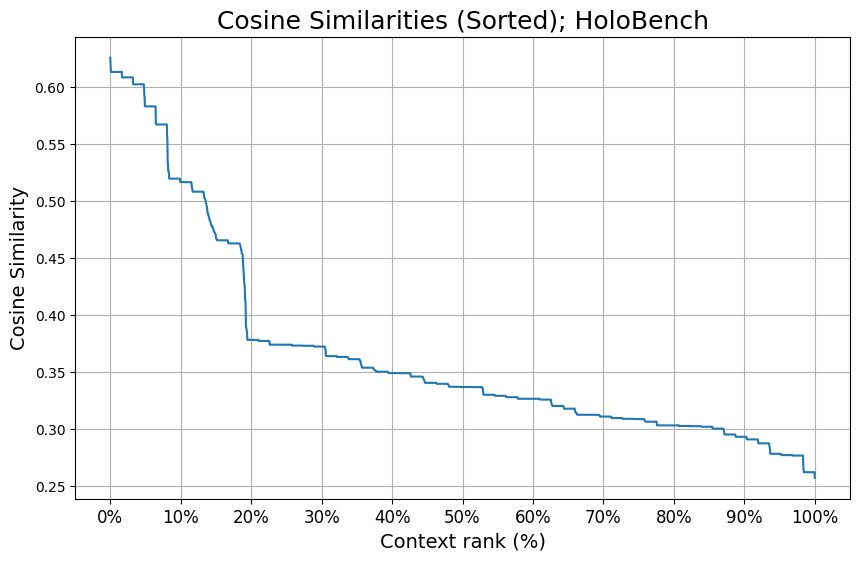}
    \caption{
        Example distributions of sorted cosine similarities from the long-context version of HotpotQA \cite{yang-etal-2018-hotpotqa} included in HELMET \cite{yen2025helmetevaluatelongcontextlanguage} with 1,000 context documents (top) and HoloBench \cite{maekawa2025holisticreasoninglongcontextlms} with 10\% relevant information amount (bottom).
        BAAI's bge-large-en-v1.5 is used as the embedding model.
    }
    \label{fig:cos-sim}
\end{figure}

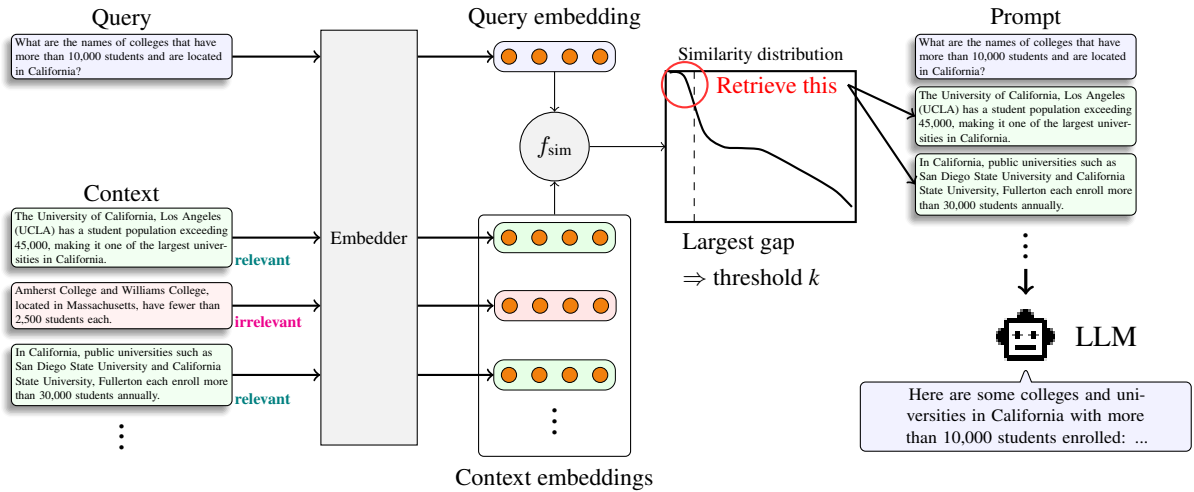
\begin{figure*}[t]
    \centering
    \input{src/workflow}
    \caption{The proposed method in the RAG workflow.
    The method chooses the threshold \textit{k} for retrieval based on a large gap in the sorted similarity score distribution.}
    \label{fig:workflow}
\end{figure*}

\subsection{Toward efficient adaptive retrieval}
\label{sec:method-toward}

\paragraph{Design motivation and principles.}

While vanilla RAG offers modularity and straightforward integration, its fixed retrieval size limits performance and efficiency in scenarios where the quantity of relevant context varies unpredictably such as in aggregation QA in the HoloBench benchmark \cite{maekawa2025holisticreasoninglongcontextlms}.
To address these limitations, we aim to design an adaptive retrieval mechanism that:
(1) operates independently of the underlying inference model and requires no additional training or fine-tuning (\textbf{Plug-and-Play}),
(2) flexibly controls the retrieval amount for each query, avoiding both wasting tokens and omitting key evidence (\textbf{Retrieval Amount Variability}), and
(3) operates in a single pass without requiring iterative LLM calls (\textbf{Single Retrieval Operation}).

\paragraph{Preliminary analysis.}
To ground our design in empirical evidence, we conduct an in-depth analysis of the distributional patterns of cosine similarity scores between queries and candidate documents, which, crucially, are inference model-agnostic signals. This preliminary analysis reveals distinct distributional characteristics that inform our adaptive retrieval strategy.

As shown in Figure~\ref{fig:cos-sim}, for factoid QA tasks such as HotpotQA, the sorted similarity scores typically exhibit a pronounced gap separating a cluster of highly relevant documents from the rest, suggesting a natural threshold for retrieval.
In contrast, aggregation tasks (e.g., HoloBench) show more irregular patterns, with gaps dispersed throughout the distribution -- reflecting the variable spread of relevant information.
In the bottom example in Figure~\ref{fig:cos-sim}, the 100k-token context is generated such that 10\% of it is information relevant to the query.
Indeed, the large gaps are observed around the top 5\% to 20\% context, aligning with our expectations.

These insights lead to the hypothesis that the largest gap in sorted similarity scores corresponds to the boundary between relevant and irrelevant documents, thus providing a data-driven criterion for adaptive retrieval size selection.

\subsection{Proposed method}

Building on these observations, we formalize an algorithm that adaptively estimates the retrieval threshold $k$ by identifying the position of the steepest drop in the similarity score distribution. The method proceeds as follows:
Compute the cosine similarities $\bm{s}$ of the query $\bm{q}$ and context documents $\bm{C}$.
Sort the scores in descending order.
Compute their first discrete differences $\bm{g}$ and choose the index $k$ where the similarity drop is the largest.
Figure~\ref{fig:workflow} depicts this process within the RAG workflow. Under the assumption that the embeddings of documents are precomputed, the time complexity of this algorithm is $\mathcal{O}(n \log n)$. 
The algorithm is described in Algorithm~\ref{alg:adaptive-k}.

\begin{algorithm}[t]
\caption{Adaptive $k$ Estimation via Largest Similarity Gap}\label{alg:adaptive-k}
\begin{algorithmic}
\Require $q$, $C$, \texttt{Embedder}($\cdot$), \texttt{Similarity}($\cdot$)
\Ensure Estimated $k$ such that the largest similarity drop occurs before the $k$-th item
\State $\bm{q} \gets \texttt{Embedder}(q)$
\State $\bm{C} \gets \texttt{Embedder}(C)$ \Comment{Precomputed}
\State $\bm{s} \gets \texttt{Similarity}(\bm{q}, \bm{C})$
\State Sort $\bm{s}$ in descending order
\State $\bm{g} \gets \texttt{array}()$ \Comment{For storing the gap}
\For{$i = 0$ to $|\bm{s}| - 2$}
    \State Append $\bm{s}[i] - \bm{s}[i+1]$ to $\bm{g}$
\EndFor
\State $k \gets \arg\max(\bm{g})$ \Comment{Index at the largest gap}
\State \Return $k$
\end{algorithmic}
\end{algorithm}

In practice, while determining the threshold $k$ based on the largest similarity gap is effective, a naïve implementation might miss relevant documents located immediately beyond the identified threshold.
To address this, we incorporate a small fixed buffer, retrieving an additional $B$ documents after the $k$-th document.
In our experiments, we set $B=5$.
Furthermore, as depicted in Figure~\ref{fig:cos-sim}, the largest gap may occasionally manifest among the least relevant documents, leading to the retrieval of an excessively large portion of the context.
To avoid this and align with our focus on retrieval from extremely long contexts, we restrict the search for the largest gap to the top 90\% of documents sorted by their similarity scores.

%% file: src/workflow.tex
\begin{tikzpicture}[
    box/.style={
        draw,
        minimum width=1.5cm,
        minimum height=0.8cm},
    vec/.style={
        draw,
        minimum width=1.3cm,
        minimum height=0.4cm},
    arrow/.style={->, thick},
]

\newcommand{\boxscale}[0]{0.4}
\newcommand{\querytextwidth}[0]{7cm}
\newcommand{\querycolor}[0]{blue!5}
\newcommand{\relevantcontextcolor}[0]{green!5}
\newcommand{\irrelevantcontextcolor}[0]{red!5}

\newcommand{\query}[0]{What are the names of colleges that have more than 10,000 students and are located in California?}
\node[box,
    scale=\boxscale,
    align=left,
    text width=\querytextwidth,
    fill=\querycolor,
    rounded corners=2pt,
    blur shadow={shadow xshift=-2pt}]
    (query) at (0, 1) {\query};
\node[align=center] at ($(query.north)+(0,0.2)$) {\small Query};

\newcommand{\relevancexshift}[0]{-0.1}
\newcommand{\contextone}[0]{The University of California, Los Angeles (UCLA) has a student population exceeding 45,000, making it one of the largest universities in California.}
\node[box,
    scale=\boxscale,
    align=left,
    text width=\querytextwidth,
    fill=\relevantcontextcolor,
    rounded corners=2pt,
    blur shadow={shadow xshift=-2pt},]
    (context1) at (0, -1.4)
    {\contextone};
\contextone
\node[align=center] at ($(context1.north)+(0,0.2)$) {\small Context};
\node[align=left, anchor=west] at ($(context1.south east) + (\relevancexshift, 0.1)$) {\tiny \color{teal} \textbf{relevant}};

\newcommand{\contexttwo}[0]{Amherst College and Williams College, located in Massachusetts, have fewer than 2,500 students each.}
\node[box,
    scale=\boxscale,
    align=left,
    text width=\querytextwidth,
    fill=\irrelevantcontextcolor,
    rounded corners=2pt,
    blur shadow={shadow xshift=-2pt},
    anchor=north]
    (context2) at ($(context1.south)+(0,-0.2)$)
    {\contexttwo};
\node[align=left, anchor=west] at ($(context2.south east) + (\relevancexshift, 0.1)$) {\tiny \color{magenta} \textbf{irrelevant}};

\newcommand{\contextthree}[0]{In California, public universities such as San Diego State University and California State University, Fullerton each enroll more than 30,000 students annually.}
\node[box,
    scale=\boxscale,
    align=left,
    text width=\querytextwidth,
    fill=\relevantcontextcolor,
    rounded corners=2pt,
    blur shadow={shadow xshift=-2pt},
    anchor=north]
    (context3) at ($(context2.south)+(0,-0.2)$)
    {\contextthree};
\node[align=left, anchor=west] at ($(context3.south east) + (\relevancexshift, 0.1)$) {\tiny \color{teal} \textbf{relevant}};

\node at ($(context3.south) + (0, -0.3)$) {\Large$\vdots$};

\node[box,
    minimum height=5.5cm,
    minimum width=0.5cm,
    fill=gray!10
    ]
    (embedder) at ($(context1.east) + (1.8, 0)$)
    {\scriptsize Embedder};

\newcommand{\embcircleradius}[0]{0.25}
\newcommand{\embcapsulescale}[0]{0.4}
\newcommand{\capsulewidth}[0]{4}
\newcommand{\capsuleheight}[0]{1}
\newcommand{\capsulerounding}[0]{10pt}

\coordinate (qvec_anchor) at ($(embedder.east |- query.center) + (1, -0.5*\embcapsulescale)$);
\coordinate (qvec_west) at ($(embedder.east |- query.center) + (1, 0)$);

\node (qvec) [inner sep=0pt, scale=\embcapsulescale] at (qvec_anchor) {
  \begin{tikzpicture}[remember picture, overlay]
    \draw[fill=\querycolor, rounded corners=\capsulerounding] (0, 0) rectangle ++(\capsulewidth, \capsuleheight);

    \foreach \i in {0.5, 1.5, 2.5, 3.5} {
      \filldraw[fill=orange!80!brown, draw=black] (\i, 0.5) circle [radius=\embcircleradius];
    }
  \end{tikzpicture}
};
\node[align=center] at ($(qvec)+(0.5*\capsulewidth*\embcapsulescale,0.7)$) {\small Query embedding};

\coordinate (cvec1_anchor) at ($(embedder.east |- context1.center) + (1, -0.5*\embcapsulescale)$);
\coordinate (cvec1_west) at ($(embedder.east |- context1.center) + (1, 0)$);

\node (cvec1) [inner sep=0pt, scale=\embcapsulescale] at (cvec1_anchor) {
  \begin{tikzpicture}[remember picture, overlay]
    \draw[fill=green!10, rounded corners=\capsulerounding] (0, 0) rectangle ++(\capsulewidth, \capsuleheight);

    \foreach \i in {0.5, 1.5, 2.5, 3.5} {
      \filldraw[fill=orange!80!brown, draw=black] (\i, 0.5) circle [radius=\embcircleradius];
    }
  \end{tikzpicture}
};

\coordinate (cvec2_anchor) at ($(embedder.east |- context2.center) + (1, -0.5*\embcapsulescale)$);
\coordinate (cvec2_west) at ($(embedder.east |- context2.center) + (1, 0)$);

\node (cvec2) [inner sep=0pt, anchor=west, scale=\embcapsulescale] at (cvec2_anchor) {
  \begin{tikzpicture}[remember picture, overlay]
    \draw[fill=red!10, rounded corners=\capsulerounding] (0, 0) rectangle ++(\capsulewidth, \capsuleheight);

    \foreach \i in {0.5, 1.5, 2.5, 3.5} {
      \filldraw[fill=orange!80!brown, draw=black] (\i, 0.5) circle [radius=\embcircleradius];
    }
  \end{tikzpicture}
};

\coordinate (cvec3_anchor) at ($(embedder.east |- context3.center) + (1, -0.5*\embcapsulescale)$);
\coordinate (cvec3_west) at ($(embedder.east |- context3.center) + (1, 0)$);

\node (cvec3) [inner sep=0pt, scale=\embcapsulescale] at (cvec3_anchor) {
  \begin{tikzpicture}[remember picture, overlay]
    \draw[fill=green!10, rounded corners=\capsulerounding] (0, 0) rectangle ++(\capsulewidth, \capsuleheight);

    \foreach \i in {0.5, 1.5, 2.5, 3.5} {
      \filldraw[fill=orange!80!brown, draw=black] (\i, 0.5) circle [radius=\embcircleradius];
    }
  \end{tikzpicture}
};

\node at ($(cvec3.south) + (0.8, -0.3)$) {\Large$\vdots$};

\node[box,
    minimum width=2cm,
    minimum height=3.2cm,
    rounded corners=2pt
    ] (vecbox)
    at ($(cvec1) + (0.8, -1.1)$) {};
\node[align=center] at ($(vecbox.south)+(0,-0.3)$) {\small Context embeddings};

\node[circle,
    draw,
    minimum size=0.5cm,
    fill=gray!10] (sim)
    at ($(vecbox.north) + (0, 0.9)$) {\small $f_\text{sim}$};

\draw[->] (vecbox.north) -- (sim.south);
\draw[->] ($(qvec) + (\capsulewidth*\embcapsulescale*0.5, 0)$) -- (sim.north);

\newcommand{\plotheight}[0]{2}
\newcommand{\plotwidth}[0]{2.5}

\draw[thick] ($(sim.east) + (1, -0.5*\plotheight)$) rectangle ++(\plotwidth,\plotheight);
\coordinate (plotwest) at ($(sim.east) + (1, 0)$);
\coordinate (plotnorth) at ($(plotwest) + (0.5*\plotwidth, 0.5*\plotheight)$);
\draw[->] (sim.east) -- (plotwest);
\node at ($(plotnorth) + (0, 0.2)$) {\scriptsize Similarity distribution};

\begin{scope}[shift={($(sim.east) + (1, -0.5*\plotheight)$)}]
    \draw[thick]
        plot [smooth] coordinates {
            (\plotwidth*0.02, \plotheight*0.99)
            (\plotwidth*0.1, \plotheight*0.96)
            (\plotwidth*0.2, \plotheight*0.6)
            (\plotwidth*0.3, \plotheight*0.5)
            (\plotwidth*0.4, \plotheight*0.49)
            (\plotwidth*0.5, \plotheight*0.48)
            (\plotwidth*0.6, \plotheight*0.42)
            (\plotwidth*0.7, \plotheight*0.35)
            (\plotwidth*0.8, \plotheight*0.28)
            (\plotwidth*0.9, \plotheight*0.20)
            (\plotwidth*0.98, \plotheight*0.10)
        };

    \draw[dashed] (\plotwidth*0.15, 0) -- (\plotwidth*0.15, \plotheight);
    \node[align=left]
        at (\plotwidth*0.15*3, -0.5)
        {\footnotesize Largest gap \\ \footnotesize $\Rightarrow$ threshold \textit{k}};

    \node[circle,
        draw=red!80,
        minimum size=0.6cm,
        thick] (retrieval)
        at (\plotwidth*0.1, \plotheight*0.91) {};
    \node[align=left] (retrievethis) at ($(retrieval.east)+(0.9, 0)$)
        {\footnotesize\color{red} Retrieve this};
\end{scope}

\node[box,
    scale=\boxscale,
    align=left,
    text width=\querytextwidth,
    fill=\querycolor,
    rounded corners=2pt,
    blur shadow={shadow xshift=-2pt}]
    (query_new) at ($(qvec |- query.east) + (7, 0)$) {\query};
\node[align=center] at ($(query_new.north)+(0,0.2)$) {\small Prompt};

\node[box,
    scale=\boxscale,
    align=left,
    text width=\querytextwidth,
    fill=\relevantcontextcolor,
    rounded corners=2pt,
    blur shadow={shadow xshift=-2pt},
    anchor=north]
    (context1_new) at ($(query_new.south) + (0,-0.1)$)
    {\contextone};

\node[box,
    scale=\boxscale,
    align=left,
    text width=\querytextwidth,
    fill=\relevantcontextcolor,
    rounded corners=2pt,
    blur shadow={shadow xshift=-2pt},
    anchor=north]
    (context3_new) at ($(context1_new.south)+(0,-0.1)$)
    {\contextthree};

\node (vdots) at ($(context3_new.south) + (0, -0.3)$) {\Large$\vdots$};

\node (llm) at ($(vdots.south) + (0, -0.9)$)
    {\includegraphics{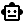}};

\node at ($(llm.east) + (0.5, 0)$) {LLM};

\node[draw,
    fill=\querycolor,
    rectangle callout,
    callout absolute pointer={($(llm.south) + (0, 0.15)$)},
    rounded corners=2pt, 
    text width=\querytextwidth*1,
    align=center,
    scale=\boxscale*1.5]
    at ($(llm.south) + (0, -0.5)$)
    {Here are some colleges and universities in California with more than 10,000 students enrolled: ...};

\draw[arrow] (query.east) -- ($ (embedder.west |- query.east) $);
\draw[arrow] (context1.east) -- ($ (embedder.west |- context1.east) $);
\draw[arrow] (context2.east) -- ($ (embedder.west |- context2.east) $);
\draw[arrow] (context3.east) -- ($ (embedder.west |- context3.east) $);
\draw[arrow] ($(embedder.east |- query.east)$) -- (qvec_west);
\draw[arrow] ($(embedder.east |- context1.east)$) -- (cvec1_west);
\draw[arrow] ($(embedder.east |- context2.east)$) -- (cvec2_west);
\draw[arrow] ($(embedder.east |- context3.east)$) -- (cvec3_west);
\draw[arrow] (retrievethis.east) -- (context1_new.west);
\draw[arrow] (retrievethis.east) -- (context3_new.west);
\draw[arrow, very thick] (vdots.south) -- (llm.north);

\end{tikzpicture}

%% file: sections/04_b_experimental.tex
\section{Experimental setup}

{
In our experiments, we aim to answer the following research questions:
\begin{itemize}
    \item How does the proposed adaptive-$k$ method compare to other modular retrieval approaches on aggregation tasks with varying amounts of relevant context?
    \item How does performance of Adaptive-$k$ vary across factoid QA and aggregation QA tasks?
    \item How does the performance gain from Adaptive‑$k$ retrieval vary across LLMs?
    \item How do different embedding models influence the performance of Adaptive-$k$?

\end{itemize}
To answer these questions, we employ the experimental settings detailed below.
}

\subsection{Dataset}
For testing on factoid QA tasks, we use HotpotQA \cite{yang-etal-2018-hotpotqa}, Natural Questions (NQ) \cite{kwiatkowski-etal-2019-natural}, and TriviaQA \cite{joshi-etal-2017-triviaqa}, as curated by HELMET \cite{yen2025helmetevaluatelongcontextlanguage} for long-context benchmarking with 128k input tokens.
Due to the high computational cost of long-context inference, we evaluate on a subset of 100 examples per dataset.

For aggregation tasks, we employ HoloBench \cite{maekawa2025holisticreasoninglongcontextlms}, which provides 90 evaluation samples.
HoloBench allows control over both total context size and the amount of information relevant to the query. We fix the total context to 100k tokens and evaluate under varying levels of relevant information, with $\texttt{info\_amount} =$ \{5000, 10000, 25000, 50000\} tokens.

\subsection{Models}

\paragraph{Retriever.}
We test our method on small, medium, and large embedding models:
Meta's contriver-msmarco\footnote{\url{https://huggingface.co/facebook/contriever-msmarco}} \citep{izacard2021contriever} with 109M params,
BAAI's bge-en-large-v1.5\footnote{\url{https://huggingface.co/BAAI/bge-large-en-v1.5}} \cite{bge_embedding} with 335M params,
and Alibaba NLP's gte-Qwen2-1.5B-instruct\footnote{\url{https://huggingface.co/Alibaba-NLP/gte-Qwen2-1.5B-instruct}} \cite{li2023towards} with 1.78B params.

\paragraph{Reader.} 
We use five closed and open models: GPT-4o-mini, GPT-4o \cite{openai2024gpt4ocard}, 
Gemini-2.5-Flash \cite{geminiteam2024geminifamilyhighlycapable},
Llama4-Scout, and
Llama4-Maverick \cite{touvron2023llamaopenefficientfoundation}.
The model details are provided in Appendix~\ref{sec:appendix-experimental}. 

\subsection{Compared methods}
We compare the proposed adaptive-$k$ method against zero-shot LLMs (without context), LLMs with full context, and \textsc{Self-Route} \cite{li-etal-2024-retrieval}, which is another modular retrieval method with a single retrieval step.
In \textsc{Self-Route}, fixed top 5k tokens are retrieved for the first inference step.
We also show the results of the fixed-$n$ retrieval method with varying numbers of tokens $n$ as performance references.
Specifically, we run experiments with $n \in$ \{1000, 5000, 10000, 25000, 50000\} and regard the best-performing setting as the oracle.
In this way, we can compare the performance of adaptive-$k$ against the best possible score of the fixed retrieval method.

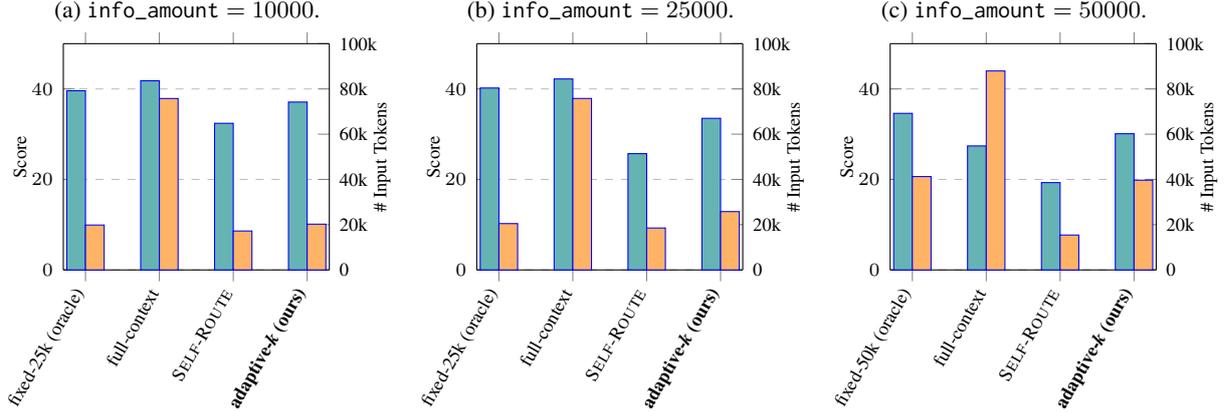
\begin{figure*}[!t]
    \centering
    \newcommand{\subfigwidth}[0]{0.32\textwidth}
    \begin{subfigure}[b]{\subfigwidth}
        \centering
        \caption{$\texttt{info\_amount} = 10000$.}
        \label{fig:holobench-results-10k}
        \input{results/holobench/bge/infos-gpt-4o/info10k}
    \end{subfigure} \hfill
    \begin{subfigure}[b]{\subfigwidth}
        \centering
        \caption{$\texttt{info\_amount} = 25000$.}
        \label{fig:holobench-results-25k}
        \input{results/holobench/bge/infos-gpt-4o/info25k}
    \end{subfigure} \hfill
    \begin{subfigure}[b]{\subfigwidth}
        \centering
        \caption{$\texttt{info\_amount} = 50000$.}
        \label{fig:holobench-results-50k}
        \input{results/holobench/bge/infos-gpt-4o/info50k}
    \end{subfigure} \\ \vspace{-1.5em}
    \caption{The results with different amounts of relevant information in the HoloBench tasks.
    The best-performing fixed-\textit{n} setting is chosen as the oracle.
    \textcolor{teal}{\rule{1ex}{2ex}} is for performance improvement, and \textcolor{orange}{\rule{1ex}{2ex}} for the number of input tokens.
    }
    \label{fig:holobench-results-info-amounts}
\end{figure*}

\subsection{Metrics} 
To evaluate the retrieval performance, context recall \cite{ru2024ragcheckerfinegrainedframeworkdiagnosing} is computed, which represents how much of the relevant context documents were able to be retrieved.
For the evaluation of generation performance, we use substring exact match (SubEM) for HotpotQA, NQ, and TriviaQA, and LLM-as-a-judge for HoloBench, following the metrics used in their original implementation in HELMET \cite{yen2025helmetevaluatelongcontextlanguage} and HoloBench, respectively.
LLM-as-a-judge evaluates whether the generated answer contains a correct mention of the gold answer, assigning a score of 1 if it finds a correct mention, 0.5 for a partially correct mention, and 0 otherwise.
For the judge model, GPT-4o-mini is used.
To evaluate the inference cost, we count the number of input and output tokens, assuming that the financial cost on the user's end and energy consumption depends on the amount of tokens \cite{husom2024pricepromptingprofilingenergy}.

%% file: results/holobench/bge/infos-gpt-4o/info10k.tex
\begin{tikzpicture}
\newcommand{\resultwidth}[0]{3.5cm}
\newcommand{\resultheight}[0]{3cm}
\pgfplotsset{set layers}
\begin{axis}[
    scale only axis,
    ybar,
    bar width=7pt,
    width=\resultwidth,
    height=\resultheight,
    ymin=0, ymax=50,
    axis y line*=left,
    ylabel={Score},
    ylabel style={yshift=-0.5em},
    symbolic x coords={fixed-25k (oracle),
        full-context,
        \textsc{Self-Route},
        \textbf{adaptive-\textit{k} (ours)}},
    xtick=data,
    xticklabel style={rotate=60, anchor=east},
    xlabel style={yshift=0.5em},
    ymajorgrids=true,
    grid style=dashed,
    tick label style={font=\scriptsize},
    label style={font=\scriptsize},
]
\addplot+[ybar, bar shift=-3.5pt, fill=teal!60] coordinates {
    (fixed-25k (oracle), 39.6)
    (full-context, 41.8)
    (\textsc{Self-Route}, 32.4)
    (\textbf{adaptive-\textit{k} (ours)}, 37.1)
};
\end{axis}
\begin{axis}[
    scale only axis,
    ybar,
    bar width=7pt,
    width=\resultwidth,
    height=\resultheight,
    ymin=0, ymax=100000,
    axis y line*=right,
    axis x line=none,
    ylabel={\# Input Tokens},
    ylabel style={yshift=.8em, font=\scriptsize},
    ytick={0,20000,40000,60000,80000,100000},
    yticklabels={0,20k,40k,60k,80k,100k},
    tick label style={font=\scriptsize},
    scaled y ticks=false,
    symbolic x coords={fixed-25k (oracle),
        full-context,
        \textsc{Self-Route},
        \textbf{adaptive-\textit{k} (ours)}},
    xtick=data,
    xticklabels={},
    legend style={at={(0.5,1.05)}, anchor=south, legend columns=-1},
]
\addplot+[ybar, bar shift=3.5pt, fill=orange!60] coordinates {
    (fixed-25k (oracle), 19839)
    (full-context, 75768)
    (\textsc{Self-Route}, 17208)
    (\textbf{adaptive-\textit{k} (ours)}, 20234)
};
\end{axis}
\end{tikzpicture}

%% file: results/holobench/bge/infos-gpt-4o/info25k.tex
\begin{tikzpicture}
\newcommand{\resultwidth}[0]{3.5cm}
\newcommand{\resultheight}[0]{3cm}
\pgfplotsset{set layers}
\begin{axis}[
    scale only axis,
    ybar,
    bar width=7pt,
    width=\resultwidth,
    height=\resultheight,
    ymin=0, ymax=50,
    axis y line*=left,
    ylabel={Score},
    ylabel style={yshift=-0.5em},
    symbolic x coords={fixed-25k (oracle),
        full-context,
        \textsc{Self-Route},
        \textbf{adaptive-\textit{k} (ours)}},
    xtick=data,
    xticklabel style={rotate=60, anchor=east},
    xlabel style={yshift=0.5em},
    ymajorgrids=true,
    grid style=dashed,
    tick label style={font=\scriptsize},
    label style={font=\scriptsize},
]
\addplot+[ybar, bar shift=-3.5pt, fill=teal!60] coordinates {
    (fixed-25k (oracle), 40.2)
    (full-context, 42.2)
    (\textsc{Self-Route}, 25.7)
    (\textbf{adaptive-\textit{k} (ours)}, 33.5)
};
\end{axis}
\begin{axis}[
    scale only axis,
    ybar,
    bar width=7pt,
    width=\resultwidth,
    height=\resultheight,
    ymin=0, ymax=100000,
    axis y line*=right,
    axis x line=none,
    ylabel={\# Input Tokens},
    ylabel style={yshift=.8em, font=\scriptsize},
    ytick={0,20000,40000,60000,80000,100000},
    yticklabels={0,20k,40k,60k,80k,100k},
    tick label style={font=\scriptsize},
    scaled y ticks=false,
    symbolic x coords={fixed-25k (oracle),
        full-context,
        \textsc{Self-Route},
        \textbf{adaptive-\textit{k} (ours)}},
    xtick=data,
    xticklabels={},
    legend style={at={(0.5,1.05)}, anchor=south, legend columns=-1},
]
\addplot+[ybar, bar shift=3.5pt, fill=orange!60] coordinates {
    (fixed-25k (oracle), 20475)
    (full-context, 75787)
    (\textsc{Self-Route}, 18525)
    (\textbf{adaptive-\textit{k} (ours)}, 25778)
};
\end{axis}
\end{tikzpicture}

%% file: results/holobench/bge/infos-gpt-4o/info50k.tex
\begin{tikzpicture}
\newcommand{\resultwidth}[0]{3.5cm}
\newcommand{\resultheight}[0]{3cm}
\pgfplotsset{set layers}
\begin{axis}[
    scale only axis,
    ybar,
    bar width=7pt,
    width=\resultwidth,
    height=\resultheight,
    ymin=0, ymax=50,
    axis y line*=left,
    ylabel={Score},
    ylabel style={yshift=-0.5em},
    symbolic x coords={fixed-50k (oracle),
        full-context,
        \textsc{Self-Route},
        \textbf{adaptive-\textit{k} (ours)}},
    xtick=data,
    xticklabel style={rotate=60, anchor=east},
    xlabel style={yshift=0.5em},
    ymajorgrids=true,
    grid style=dashed,
    tick label style={font=\scriptsize},
    label style={font=\scriptsize},
]
\addplot+[ybar, bar shift=-3.5pt, fill=teal!60] coordinates {
    (fixed-50k (oracle), 34.6)
    (full-context, 27.4)
    (\textsc{Self-Route}, 19.3)
    (\textbf{adaptive-\textit{k} (ours)}, 30.1)
};
\end{axis}
\begin{axis}[
    scale only axis,
    ybar,
    bar width=7pt,
    width=\resultwidth,
    height=\resultheight,
    ymin=0, ymax=100000,
    axis y line*=right,
    axis x line=none,
    ylabel={\# Input Tokens},
    ylabel style={yshift=.8em, font=\scriptsize},
    ytick={0,20000,40000,60000,80000,100000},
    yticklabels={0,20k,40k,60k,80k,100k},
    tick label style={font=\scriptsize},
    scaled y ticks=false,
    symbolic x coords={fixed-50k (oracle),
        full-context,
        \textsc{Self-Route},
        \textbf{adaptive-\textit{k} (ours)}},
    xtick=data,
    xticklabels={},
    legend style={at={(0.5,1.05)}, anchor=south, legend columns=-1},
]
\addplot+[ybar, bar shift=3.5pt, fill=orange!60] coordinates {
    (fixed-50k (oracle), 41265)
    (full-context, 87936)
    (\textsc{Self-Route}, 15426)
    (\textbf{adaptive-\textit{k} (ours)}, 39654)
};
\end{axis}
\end{tikzpicture}

%% file: sections/05_results.tex
\section{Results}
This section provides the results of the experiments with a focus on different task types, reader models, and embedding models.
For the full results, see Appendix~\ref{sec:appendix-full-results}.

\subsection{Aggregation-type QA}
Figure~\ref{fig:holobench-results-info-amounts} shows GPT-4o's results in the HoloBench tasks where each task is designed to contain different amounts of relevant information (\texttt{info\_amount}: 10k, 25k, 50k tokens) in the context.
It can be observed that our Adaptive-$k$ method constantly outperforms \textsc{Self-Route}. 
The performance improvements of Adaptive-$k$ are particularly notable when the amount of relevant information in the context is high.
Also, our method flexibly increases the amount of retrieved context chunks when there is a higher amount of relevant information in the entire context.
In contrast, \textsc{Self-Route} tends to underestimate the amount of relevant context and jump to a conclusion that the LLM can answer the query with the 5k-token context retrieved in the first round, leading to lower performance in a high amount of relevant information.

This contrast is also reflected in the context recall scores.
As shown in Table~\ref{tab:holobench-results-context-recall}, Adaptive-$k$ consistently achieves a context recall score of approximately 70 across varying levels of relevant information, indicating that it retrieves approximately 70\% of truly relevant chunks regardless of their proportion in the full context.
The contrast is even more pronounced when compared to context recall of \textsc{Self-Route}, with Adaptive-$k$ achieving more than three times higher context recall.

\begin{table}[t]
    \centering
    \small
    \setlength{\tabcolsep}{5pt}
    \begin{tabular}{lrrrr} \toprule
         & info5k & info10k & info25k & info50k \\ \midrule
        \textsc{Self-Route} & 65.79 & 45.04 & 30.42 & 21.54 \\
        \textbf{Adaptive-$k$} & \textbf{75.74} & \textbf{68.54} & \textbf{66.16} & \textbf{67.43} \\ \midrule
        fixed-1k            & 12.05 & 6.53 & 2.77 & 1.47 \\
        fixed-5k            & 51.92 & 31.77 & 14.06 & 7.54 \\
        fixed-10k           & 66.68 & 59.10 & 28.80 & 15.39 \\
        fixed-25k           & 78.48 & 78.18 & 68.13 & 39.55 \\
        fixed-50k           & 86.79 & 87.34 & 86.88 & 76.90 \\
        \bottomrule
    \end{tabular}
    \caption{A comparison of the context recall scores across different relevant information amounts in the HoloBench tasks.
    The query and contexts are embedded by bge-large-en-v1.5.
    The scores compared are \textsc{Self-Route} and Adaptive-$k$, as well as the results of fixed-$n$ token retrieval as references.}
    \label{tab:holobench-results-context-recall}
\end{table}

\subsection{Factoid-type QA}

\begin{figure}
    \centering
    \input{results/summary-results_qa} \\ \vspace{-1.3em}
    \caption{
        A performance comparison of our proposed method (\textbf{Adaptive-$k$}) in the factoid QA tasks against existing methods.
        The embedding model is bge-large-en-v1.5, and the reader model is GPT-4o.
        \textcolor{teal}{\rule{1ex}{2ex}} is for the SubEM scores, and \textcolor{orange}{\rule{1ex}{2ex}} for the number of input tokens. 
    }
    \label{fig:summary-results_qa}
\end{figure}
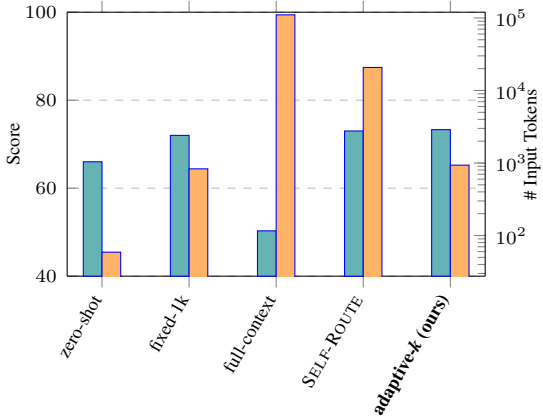

Figure~\ref{fig:summary-results_qa} shows the comparison of Adaptive-$k$ against the zero-shot setting, fixed 1k-token retrieval, full context, and \textsc{Self-Route}. 
All methods are implemented using GPT-4o.
Our method achieves a 99\% reduction in input cost compared to the full context input, and a 90\% reduction compared to \textsc{Self-Route}.
Since users generally lack prior knowledge of the optimal retrieval size, Adaptive-$k$ successfully reduces the cost while improving the generation quality compared to zero-shot question answering.

\subsection{Comparison across LLMs}

\begin{figure*}[t]
    \centering
    \newcommand{\subfigwidth}[0]{0.32\textwidth}
    \newcommand{\subfigposition}[0]{t}
    \vspace{-0.4em}
    \begin{subfigure}[\subfigposition]{\subfigwidth}
        \centering
        \caption{GPT-4o-mini.}
        \label{fig:holobench-results-main-gpt-4o-mini}
        \input{results/holobench/bge/gpt-4o-mini}
    \end{subfigure} \hfill
    \begin{subfigure}[\subfigposition]{\subfigwidth}
        \centering
        \caption{GPT-4o.}
        \label{fig:holobench-results-main-gpt-4o}
        \input{results/holobench/bge/gpt-4o}
    \end{subfigure} \hfill
    \begin{subfigure}[\subfigposition]{\subfigwidth}
        \centering
        \caption{Gemini-2.5-Flash.}
        \label{fig:holobench-results-main-gemini}
        \input{results/holobench/bge/gemini-2.5-flash}
    \end{subfigure} \\ \vspace{-1.6em}
    \begin{subfigure}[\subfigposition]{\subfigwidth}
        \centering
        \caption{Llama4-Scout.}
        \label{fig:holobench-results-main-llama4-scout}
        \input{results/holobench/bge/llama4-scout}
    \end{subfigure} \hspace{3em}
    \begin{subfigure}[\subfigposition]{\subfigwidth}
        \centering
        \caption{Llama4-Maverick.}
        \label{fig:holobench-results-main-llama4-maverick}
        \input{results/holobench/bge/llama4-maverick}
    \end{subfigure} \\ \vspace{-1.6em}
    \caption{A performance comparison across the different reader models in the HoloBench task.
    The emnbedding model is bge-large-en-v1.5.
    \textcolor{teal}{\rule{1ex}{2ex}} is for performance improvement, and \textcolor{orange}{\rule{1ex}{2ex}} for the number of input tokens.
    }
    \label{fig:holobench-results-main}
\end{figure*}
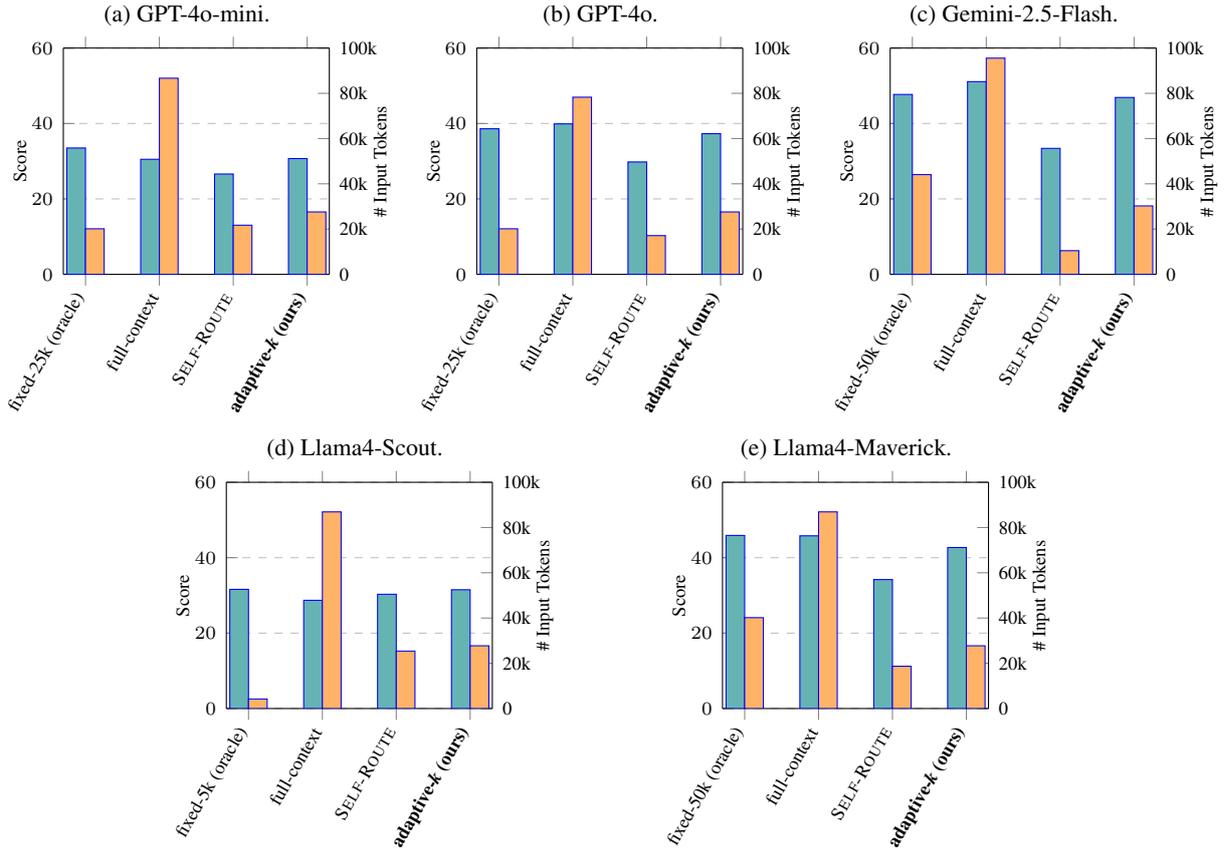

Since our methods only modify the retriever module, the retrieved documents to be fed into an LLM's prompt remain the same across different LLMs.
However, we observe that its effectiveness varies notably by model. Figure~\ref{fig:holobench-results-main} shows the average score improvements and input token counts across different relevant information settings in HoloBench. 
Larger high-performance LLMs such as
GPT-4o (Figure~\ref{fig:holobench-results-main-gpt-4o}),
Gemini-2.5-Flash (Figure~\ref{fig:holobench-results-main-gemini}),
and Llama4-Maverick (Figure~\ref{fig:holobench-results-main-llama4-maverick})
show substantial gains from Adaptive-$k$ retrieval compared to \textsc{Self-Route}.  In contrast, smaller LLMs such as
GPT-4o-mini (Figure~\ref{fig:holobench-results-main-gpt-4o-mini}) and
Llama4-Scout (Figure~\ref{fig:holobench-results-main-llama4-scout}) exhibit more modest improvements. Nonetheless, even for smaller models, Adaptive-$k$ effectively reduces context length while maintaining performance close to the full-context and oracle fixed-$n$ baselines.

\subsection{Embedding bottleneck}

\input{src/embedding-bottleneck}

We observed that the effectiveness of our adaptive method is sensitive to the choice of embedding model. 
As shown in Table~\ref{tab:results-embedding-context-recall}, the embeddings by bge-large-en-v1.5, gte-Qwen2-1.5B-instruct, and contriever-msmarco have different strengths depending on the task.
In factoid QA tasks, BGE embeddings consistently yield higher context recall than GTE, whereas GTE performs better on HoloBench. 
The underlying cause remains unclear, but we identify a few potential factors: (1) Context chunk length: the factoid QA tasks in HELMET generally have a longer context chunk length (up to $\sim$100 tokens) than HoloBench ($\sim$40 tokens); (2) Chunking scheme \cite{zhong2025mixofgranularityoptimizechunkinggranularity}: while the context chunks in HoloBench contain well-formed natural-language sentences, those in the factoid QA tasks often contain mid-sentence breaks;
(3) Training scheme: differences in pretraining corpora and formatting may lead to divergent performance across embedding models. 
Overall, choosing the right embedding model is critical for ensuring RAG effectiveness. For general use, we recommend bge-large-en-v1.5 for Adaptive-$k$ due to its strong and consistent performance across settings.

\subsection{Limitation of fixed retrieval}
While fixed-$n$ retrieval occasionally outperforms Adaptive-$k$ method, it requires prior knowledge of the optimal $n$, which is difficult to estimate in practice. Our results show that the best-performing $n$ varies across task types, query types, embedding models, reader models, and the distribution of relevant information. 
In contrast, Adaptive-$k$ is able to dynamically adjust the retrieval amount based on the query and context chunks, eliminating the need for manual tuning. 
This not only removes the burden and risk of heuristically selecting an $n$ but also provides a more robust and generalizable solution across a wide range of scenarios, especially in cases where the relevant context size is highly variable or unknown a priori.

In addition, Tables~\ref{tab:diff-k-holobench} and \ref{tab:diff-k-hotpotqa} demonstrate that the Adaptive-$k$ method effectively estimates the threshold of relevant contexts across different relevant context sizes.
These tables report the average absolute difference between the estimated threshold ($k$) and the true $k$-value, referred to as diff-$k$ henceforth, across various datasets and embeddings.
The true $k$-value is defined as the index of the last relevant context chunk in the list of context chunk embeddings sorted by similarity; in other words, diff-$k$ is the smallest value of $k$ that achieves 100\% recall.
Table~\ref{tab:diff-k-holobench} presents the diff-$k$ values for the HoloBench task with 50k tokens of relevant context.
The diff-$k$ values for Adaptive-$k$ are substantially lower than those of the \textsc{Self-Route} baseline and closely match those of the fixed 50k-token retrieval, which is treated as an oracle.
Similarly, the diff-$k$ values for Adaptive-$k$ in the HotpotQA task (Table~\ref{tab:diff-k-hotpotqa}) consistently approximate the oracle retrieval results, indicating that the method reliably determines the optimal retrieval amount.

\begin{table}[t]
    \centering
    \small
    \begin{tabular}{lrrr} \toprule
        & Contriever & BGE & GTE \\ \midrule
        fixed-1k & 2351.96 & 2289.33 & 2265.12 \\
        fixed-5k & 2236.31 & 2186.08 & 2154.81 \\
        fixed-10k & 2095.38 & 2056.89 & 2017.09 \\
        fixed-25k & 1680.93 & 1660.64 & 1599.99 \\
        fixed-50k (oracle) & 992.47 & 985.39 & 898.34 \\ \midrule
        Adaptive-$k$ & \textbf{1133.28} & \textbf{1248.99} & \textbf{1001.00} \\
        \textsc{Self-Route} & 2236.31 & 2186.09 & 2154.81 \\
        \bottomrule
    \end{tabular}
    \caption{Diff-$k$ values in the HoloBench task with an information amount of 50k for the three embedding models.}
    \label{tab:diff-k-holobench}
\end{table}

\begin{table}[t]
    \centering
    \small
    \begin{tabular}{lrrr} \toprule
        & Contriever & BGE & GTE \\ \midrule
        fixed-1k (oracle) & 162.52 & 88.86 & 631.90 \\
        fixed-5k & 154.68 & 97.89 & 598.70 \\
        fixed-10k & 156.29 & 116.35 & 558.69 \\
        fixed-25k & 191.14 & 185.70 & 452.32 \\
        fixed-50k & 297.41 & 320.50 & 307.12 \\ \midrule
        Adaptive-$k$ & 161.14 & \textbf{89.40} & 632.28 \\
        \textsc{Self-Route} & \textbf{154.68} & 97.89 & 598.70 \\
        \bottomrule
    \end{tabular}
    \caption{Diff-$k$ values in HotpotQA for the three embedding models.
    Given the low retrieval performance with the GTE embeddings in the factoid-type QA tasks in Table~\ref{tab:results-embedding-context-recall}, the results of GTE embeddings in this table are not informative.}
    \label{tab:diff-k-hotpotqa}
\end{table}

%% file: results/summary-results_qa.tex
\begin{tikzpicture}
\newcommand{\resultwidth}[0]{5.5cm}
\newcommand{\resultheight}[0]{3.5cm}
\pgfplotsset{set layers}
\begin{axis}[
    scale only axis,
    ybar,
    bar width=7pt,
    width=\resultwidth,
    height=\resultheight,
    ymin=40, ymax=100,
    axis y line*=left,
    ylabel={Score},
    ylabel style={yshift=-0.5em},
    symbolic x coords={zero-shot,
        fixed-1k,
        full-context,
        \textsc{Self-Route},
        \textbf{adaptive-\textit{k} (ours)}},
    xtick=data,
    xticklabel style={rotate=60, anchor=east},
    xlabel style={yshift=0.5em},
    ymajorgrids=true,
    grid style=dashed,
    tick label style={font=\scriptsize},
    label style={font=\scriptsize},
]
\addplot+[ybar, bar shift=-3.5pt, fill=teal!60] coordinates {
    (zero-shot, 66.0)
    (fixed-1k, 72.0)
    (full-context, 50.3)
    (\textsc{Self-Route}, 73.0)
    (\textbf{adaptive-\textit{k} (ours)}, 73.3)
};
\end{axis}
\begin{axis}[
    scale only axis,
    ybar,
    bar width=7pt,
    width=\resultwidth,
    height=\resultheight,
    ymin=0, ymax=120000,
    axis y line*=right,
    axis x line=none,
    ylabel={\# Input Tokens},
    ylabel style={yshift=.8em, font=\scriptsize},
    yticklabel style={font=\scriptsize},
    tick label style={font=\scriptsize},
    ymode=log,
    log basis y=10,
    yticklabel style={/pgf/number format/fixed, /pgf/number format/precision=0},
    symbolic x coords={zero-shot,
        fixed-1k,
        full-context,
        \textsc{Self-Route},
        \textbf{adaptive-\textit{k} (ours)}},
    xtick=data,
    xticklabels={},
    legend style={at={(0.5,1.05)}, anchor=south, legend columns=-1},
]
\addplot+[ybar, bar shift=3.5pt, fill=orange!60] coordinates {
    (zero-shot, 59)
    (fixed-1k, 831)
    (full-context, 110336)
    (\textsc{Self-Route}, 20759)
    (\textbf{adaptive-\textit{k} (ours)}, 934)
};
\end{axis}
\end{tikzpicture}

%% file: results/holobench/bge/gpt-4o-mini.tex
\begin{tikzpicture}
\newcommand{\resultwidth}[0]{3.5cm}
\newcommand{\resultheight}[0]{3cm}
\pgfplotsset{set layers}
\begin{axis}[
    scale only axis,
    ybar,
    bar width=7pt,
    width=\resultwidth,
    height=\resultheight,
    ymin=0, ymax=60,
    axis y line*=left,
    ylabel={Score},
    ylabel style={yshift=-0.5em},
    symbolic x coords={fixed-25k (oracle),
        full-context,
        \textsc{Self-Route},
        \textbf{adaptive-\textit{k} (ours)}},
    xtick=data,
    xticklabel style={rotate=60, anchor=east},
    xlabel style={yshift=0.5em},
    ymajorgrids=true,
    grid style=dashed,
    tick label style={font=\scriptsize},
    label style={font=\scriptsize},
]
\addplot+[ybar, bar shift=-3.5pt, fill=teal!60] coordinates {
    (fixed-25k (oracle), 33.5)
    (full-context, 30.5)
    (\textsc{Self-Route}, 26.6)
    (\textbf{adaptive-\textit{k} (ours)}, 30.7)
};
\end{axis}
\begin{axis}[
    scale only axis,
    ybar,
    bar width=7pt,
    width=\resultwidth,
    height=\resultheight,
    ymin=0, ymax=100000,
    axis y line*=right,
    axis x line=none,
    ylabel={\# Input Tokens},
    ylabel style={yshift=.8em, font=\scriptsize},
    ytick={0,20000,40000,60000,80000,100000},
    yticklabels={0,20k,40k,60k,80k,100k},
    tick label style={font=\scriptsize},
    scaled y ticks=false,
    symbolic x coords={fixed-25k (oracle),
        full-context,
        \textsc{Self-Route},
        \textbf{adaptive-\textit{k} (ours)}},
    xtick=data,
    xticklabels={},
    legend style={at={(0.5,1.05)}, anchor=south, legend columns=-1},
]
\addplot+[ybar, bar shift=3.5pt, fill=orange!60] coordinates {
    (fixed-25k (oracle), 20128)
    (full-context, 86678)
    (\textsc{Self-Route}, 21711)
    (\textbf{adaptive-\textit{k} (ours)}, 27573)
};
\end{axis}
\end{tikzpicture}

%% file: results/holobench/bge/gpt-4o.tex
\begin{tikzpicture}
\newcommand{\resultwidth}[0]{3.5cm}
\newcommand{\resultheight}[0]{3cm}
\pgfplotsset{set layers}
\begin{axis}[
    scale only axis,
    ybar,
    bar width=7pt,
    width=\resultwidth,
    height=\resultheight,
    ymin=0, ymax=60,
    axis y line*=left,
    ylabel={Score},
    ylabel style={yshift=-0.5em},
    symbolic x coords={fixed-25k (oracle),
        full-context,
        \textsc{Self-Route},
        \textbf{adaptive-\textit{k} (ours)}},
    xtick=data,
    xticklabel style={rotate=60, anchor=east},
    xlabel style={yshift=0.5em},
    ymajorgrids=true,
    grid style=dashed,
    tick label style={font=\scriptsize},
    label style={font=\scriptsize},
]
\addplot+[ybar, bar shift=-3.5pt, fill=teal!60] coordinates {
    (fixed-25k (oracle), 38.6)
    (full-context, 39.9)
    (\textsc{Self-Route}, 29.8)
    (\textbf{adaptive-\textit{k} (ours)}, 37.3)
};
\end{axis}
\begin{axis}[
    scale only axis,
    ybar,
    bar width=7pt,
    width=\resultwidth,
    height=\resultheight,
    ymin=0, ymax=100000,
    axis y line*=right,
    axis x line=none,
    ylabel={\# Input Tokens},
    ylabel style={yshift=.8em, font=\scriptsize},
    ytick={0,20000,40000,60000,80000,100000},
    yticklabels={0,20k,40k,60k,80k,100k},
    tick label style={font=\scriptsize},
    scaled y ticks=false,
    symbolic x coords={fixed-25k (oracle),
        full-context,
        \textsc{Self-Route},
        \textbf{adaptive-\textit{k} (ours)}},
    xtick=data,
    xticklabels={},
    legend style={at={(0.5,1.05)}, anchor=south, legend columns=-1},
]
\addplot+[ybar, bar shift=3.5pt, fill=orange!60] coordinates {
    (fixed-25k (oracle), 20128)
    (full-context, 78286)
    (\textsc{Self-Route}, 17105)
    (\textbf{adaptive-\textit{k} (ours)}, 27573)
};
\end{axis}
\end{tikzpicture}

%% file: results/holobench/bge/gemini-2.5-flash.tex
\begin{tikzpicture}
\newcommand{\resultwidth}[0]{3.5cm}
\newcommand{\resultheight}[0]{3cm}
\pgfplotsset{set layers}
\begin{axis}[
    scale only axis,
    ybar,
    bar width=7pt,
    width=\resultwidth,
    height=\resultheight,
    ymin=0, ymax=60,
    axis y line*=left,
    ylabel={Score},
    ylabel style={yshift=-0.5em},
    symbolic x coords={fixed-50k (oracle),
        full-context,
        \textsc{Self-Route},
        \textbf{adaptive-\textit{k} (ours)}},
    xtick=data,
    xticklabel style={rotate=60, anchor=east},
    xlabel style={yshift=0.5em},
    ymajorgrids=true,
    grid style=dashed,
    tick label style={font=\scriptsize},
    label style={font=\scriptsize},
]
\addplot+[ybar, bar shift=-3.5pt, fill=teal!60] coordinates {
    (fixed-50k (oracle), 47.7)
    (full-context, 51.1)
    (\textsc{Self-Route}, 33.4)
    (\textbf{adaptive-\textit{k} (ours)}, 46.9)
};
\end{axis}
\begin{axis}[
    scale only axis,
    ybar,
    bar width=7pt,
    width=\resultwidth,
    height=\resultheight,
    ymin=0, ymax=100000,
    axis y line*=right,
    axis x line=none,
    ylabel={\# Input Tokens},
    ylabel style={yshift=.8em, font=\scriptsize},
    ytick={0,20000,40000,60000,80000,100000},
    yticklabels={0,20k,40k,60k,80k,100k},
    tick label style={font=\scriptsize},
    scaled y ticks=false,
    symbolic x coords={fixed-50k (oracle),
        full-context,
        \textsc{Self-Route},
        \textbf{adaptive-\textit{k} (ours)}},
    xtick=data,
    xticklabels={},
    legend style={at={(0.5,1.05)}, anchor=south, legend columns=-1},
]
\addplot+[ybar, bar shift=3.5pt, fill=orange!60] coordinates {
    (fixed-50k (oracle), 44071)
    (full-context, 95542)
    (\textsc{Self-Route}, 10450)
    (\textbf{adaptive-\textit{k} (ours)}, 30235)
};
\end{axis}
\end{tikzpicture}

%% file: results/holobench/bge/llama4-scout.tex
\begin{tikzpicture}
\newcommand{\resultwidth}[0]{3.5cm}
\newcommand{\resultheight}[0]{3cm}
\pgfplotsset{set layers}
\begin{axis}[
    scale only axis,
    ybar,
    bar width=7pt,
    width=\resultwidth,
    height=\resultheight,
    ymin=0, ymax=60,
    axis y line*=left,
    ylabel={Score},
    ylabel style={yshift=-0.5em},
    symbolic x coords={fixed-5k (oracle),
        full-context,
        \textsc{Self-Route},
        \textbf{adaptive-\textit{k} (ours)}},
    xtick=data,
    xticklabel style={rotate=60, anchor=east},
    xlabel style={yshift=0.5em},
    ymajorgrids=true,
    grid style=dashed,
    tick label style={font=\scriptsize},
    label style={font=\scriptsize},
]
\addplot+[ybar, bar shift=-3.5pt, fill=teal!60] coordinates {
    (fixed-5k (oracle), 31.6)
    (full-context, 28.7)
    (\textsc{Self-Route}, 30.3)
    (\textbf{adaptive-\textit{k} (ours)}, 31.5)
};
\end{axis}
\begin{axis}[
    scale only axis,
    ybar,
    bar width=7pt,
    width=\resultwidth,
    height=\resultheight,
    ymin=0, ymax=100000,
    axis y line*=right,
    axis x line=none,
    ylabel={\# Input Tokens},
    ylabel style={yshift=.8em, font=\scriptsize},
    ytick={0,20000,40000,60000,80000,100000},
    yticklabels={0,20k,40k,60k,80k,100k},
    tick label style={font=\scriptsize},
    scaled y ticks=false,
    symbolic x coords={fixed-5k (oracle),
        full-context,
        \textsc{Self-Route},
        \textbf{adaptive-\textit{k} (ours)}},
    xtick=data,
    xticklabels={},
    legend style={at={(0.5,1.05)}, anchor=south, legend columns=-1},
]
\addplot+[ybar, bar shift=3.5pt, fill=orange!60] coordinates {
    (fixed-5k (oracle), 4226)
    (full-context, 86909)
    (\textsc{Self-Route}, 25390)
    (\textbf{adaptive-\textit{k} (ours)}, 27729)
};
\end{axis}
\end{tikzpicture}

%% file: results/holobench/bge/llama4-maverick.tex
\begin{tikzpicture}
\newcommand{\resultwidth}[0]{3.5cm}
\newcommand{\resultheight}[0]{3cm}
\pgfplotsset{set layers}
\begin{axis}[
    scale only axis,
    ybar,
    bar width=7pt,
    width=\resultwidth,
    height=\resultheight,
    ymin=0, ymax=60,
    axis y line*=left,
    ylabel={Score},
    ylabel style={yshift=-0.5em},
    symbolic x coords={fixed-50k (oracle),
        full-context,
        \textsc{Self-Route},
        \textbf{adaptive-\textit{k} (ours)}},
    xtick=data,
    xticklabel style={rotate=60, anchor=east},
    xlabel style={yshift=0.5em},
    ymajorgrids=true,
    grid style=dashed,
    tick label style={font=\scriptsize},
    label style={font=\scriptsize},
]
\addplot+[ybar, bar shift=-3.5pt, fill=teal!60] coordinates {
    (fixed-50k (oracle), 45.9)
    (full-context, 45.8)
    (\textsc{Self-Route}, 34.2)
    (\textbf{adaptive-\textit{k} (ours)}, 42.7)
};
\end{axis}
\begin{axis}[
    scale only axis,
    ybar,
    bar width=7pt,
    width=\resultwidth,
    height=\resultheight,
    ymin=0, ymax=100000,
    axis y line*=right,
    axis x line=none,
    ylabel={\# Input Tokens},
    ylabel style={yshift=.8em, font=\scriptsize},
    ytick={0,20000,40000,60000,80000,100000},
    yticklabels={0,20k,40k,60k,80k,100k},
    tick label style={font=\scriptsize},
    scaled y ticks=false,
    symbolic x coords={fixed-50k (oracle),
        full-context,
        \textsc{Self-Route},
        \textbf{adaptive-\textit{k} (ours)}},
    xtick=data,
    xticklabels={},
    legend style={at={(0.5,1.05)}, anchor=south, legend columns=-1},
]
\addplot+[ybar, bar shift=3.5pt, fill=orange!60] coordinates {
    (fixed-50k (oracle), 40160)
    (full-context, 86909)
    (\textsc{Self-Route}, 18697)
    (\textbf{adaptive-\textit{k} (ours)}, 27729)
};
\end{axis}
\end{tikzpicture}

%% file: src/embedding-bottleneck.tex
\begin{table}[t]
    \centering
    \small
    \setlength{\tabcolsep}{4pt}
    \newcommand{\qacolwidth}[0]{6.5em}
    \begin{tabular}{@{}llrrr@{}} \toprule
         & Method & Contriever & BGE & GTE \\ \midrule
        \multirow{2}{\qacolwidth}{HotpotQA} & \textsc{Self-Route} & 77.50 & 90.83 & 25.83 \\
        & adaptive-\textit{k} & 49.17 & 70.83 & 5.50 \\ \midrule
        \multirow{2}{\qacolwidth}{NQ} & \textsc{Self-Route} & 51.19 & 51.90 & 20.67 \\
        & adaptive-\textit{k} & 26.28 & 27.20 & 2.85 \\ \midrule
        \multirow{2}{\qacolwidth}{TriviaQA} & \textsc{Self-Route} & 41.49 & 46.52 & 10.20 \\
        & adaptive-\textit{k} & 25.02 & 31.21 & 3.00 \\ \midrule\midrule
        \multirow{2}{\qacolwidth}{HoloBench-5k} & \textsc{Self-Route} & 60.57 & 65.79 & 65.18 \\
        & adaptive-\textit{k} & 72.11 & 75.74 & 82.20 \\ \midrule
        \multirow{2}{\qacolwidth}{HoloBench-10k} & \textsc{Self-Route} & 41.83 & 45.04 & 45.87 \\
        & adaptive-\textit{k} & 70.85 & 68.54 & 78.99 \\ \midrule
        \multirow{2}{\qacolwidth}{HoloBench-25k} & \textsc{Self-Route} & 29.67 & 30.42 & 31.02 \\
        & adaptive-\textit{k} & 68.67 & 66.16 & 76.47 \\ \midrule
        \multirow{2}{\qacolwidth}{HoloBench-50k} & \textsc{Self-Route} & 21.27 & 21.54 & 21.90 \\
        & adaptive-\textit{k} & 67.16 & 67.43 & 72.54 \\
        \bottomrule
    \end{tabular}
    \caption{A comparison of the context recall scores across tasks between Contriever (contriever-msmarco), BGE (bge-large-en-v1.5), and GTE (gte-Qwen2-1.5B-instruct).}
    \label{tab:results-embedding-context-recall}
\end{table}

%% file: sections/06_conclusion.tex
\section{Conclusion}
We presented a simple yet effective and efficient plug-and-play method, \textbf{adaptive-$k$}, that dynamically selects the number of context chunks to retrieve in a single step, based on the similarity distribution between the query and context chunks.
Unlike existing adaptive retrieval methods that requires iterative inference steps, our method only requires a single matrix calculation to estimate the retrieval threshold, achieving a fast and flexible retrieval module.
This method is particularly effective for aggregation-type QA tasks, where the optimal number of context chunks varies across examples and cannot be predetermined by a fixed-token retrieval strategy.
Results on HoloBench demonstrate that Adaptive-$k$ flexibly adjusts retrieval size to align with the amount of relevant information in the context.
In factoid QA tasks, where relevant information is sparse, our method aggressively prunes the context while still outperforming zero-shot QA in answer quality. 
Compared to \textsc{Self-Route}, our method consistently achieves superior performance in aggregation-type QA tasks, while drastically reducing the input size and maintaining higher context recall.

Our adaptive-$k$ retrieval is a plug-and-play, single-pass alternative to fixed-size retrieval, yet several directions remain.
First, because the method is orthogonal to most RAG pipelines, pairing it with techniques such as query-expansion, iterative reranking, or generative feedback loops could further improve accuracy and latency.
Second, embedding models excel on different query and corpus traits; a runtime system that selects or ensembles embeddings per query may unlock extra gains in recall and robustness.

%% file: sections/07_limitations.tex
\section*{Limitations}
While our proposed method shows promising results in adaptive retrieval for question answering tasks, it has several limitations that warrant discussion.

First, the method is not directly applicable to tasks such as summarization, where the objective is to process the entire input holistically rather than retrieve a subset of relevant context.
In such cases, aggressive filtering may omit important information that contributes to the overall summary.
In addition, an embedding model is not able to identify the relevant context documents with a general summarization-type query.
For instance, when the query for a summarization task is a general statement like \texttt{``The summary of this book is:''} (an example from $\infty$\textsc{Bench} Sum \cite{zhang2024inftybenchextendinglongcontext}), the high-similarity context chunks do not necessarily reflect the importance to the answer because the query does not quite contain semantically significant information.

Second, our method is designed for natural language inputs and assumes meaningful semantic similarity between queries and context chunks.
It does not generalize well to non-natural-language tasks, such as those involving structured key-value formats (\textit{e.g.}, JSON), where semantic embeddings may not capture relevance effectively.

Third, the approach is sensitive to surface-level variations in text.
For example, typographical errors in the query or context can negatively affect embedding quality and distort similarity scores, leading to suboptimal retrieval decisions.
If the queries are expected to be noisy with non-standard spellings or grammar, adding a query standardization module \cite{chan2024rqraglearningrefinequeries} on top of our adaptive-$k$ method would be helpful.

Lastly, the method may be vulnerable to adversarial or malicious inputs \cite{wallace-etal-2019-universal}.
A specially crafted context chunk could receive an artificially high or low similarity score, thereby introducing a large gap in the similarity distribution and misleading the algorithm into selecting an incorrect retrieval threshold \cite{su2024robustretrievalaugmentedgenerationevaluating}.
Mitigating such risks would require additional robustness checks or adversarial training techniques, which are beyond the scope of this work.

%% file: sections/08_ethical.tex
\section*{Ethical considerations}
One of the key advantages of our proposed adaptive retrieval method is its potential to reduce the environmental impact of LLM inference.
By discarding irrelevant context chunks and only retrieving a minimal yet sufficient subset of documents, our approach significantly reduces the number of input tokens processed.
In our experiments, our proposed method discarded nearly 99\% of the input tokens in factoid QA tasks, and substantially reduced input size in aggregation QA tasks while maintaining high context recall.

This reduction translates into lower computational overhead, leading to more energy-efficient inference.
As a result, our method contributes to decreasing the carbon footprint associated with deploying LLMs at scale.
With the growing trend of longer context windows, flexibly filtering out irrelevant context is necessary to ensure energy-efficient inference.

While efficiency is a central goal, we emphasize that any optimization must not compromise fairness or content coverage.
Our method is designed to be model-agnostic and does not introduce or amplify biases beyond those present in the similarity scoring mechanism, \textit{e.g.}, cosine similarity over embedding spaces.
However, care should be taken when applying this method in high-stakes domains, \textit{e.g.}, medical or legal QA, where discarding seemingly low-similarity context could result in the omission of critical information. 
Further research is needed to quantify such risks and guide responsible deployment.

While we used AI assitants such as ChatGPT and Copilot to assist in coding and revising this paper, we carefully reviewed and edited all content to ensure it meets our standards and aligns with our research goals.

%% file: sections/10_appendix.tex
\clearpage
\onecolumn
\appendix

\section{Appendix}
\subsection{Prompt templates}
\label{sec:appendix-prompt}
\subsubsection{Prompt template for the factoid QA tasks}

\begin{tcblisting}{listing only}
Your task is to answer the question provided. To help you answer accurately, some relevant
context documents have been retrieved. After reviewing them, you'll be asked the same question
again. Please respond succinctly.

**Input:**
- **Question:**
```
{question}
```

- **Context:**
```
{context}
```

- **Question:**
```
{question}
```

**Response:**
- **Answer:**
\end{tcblisting}

\subsubsection{Prompt template for the HoloBench tasks}
\begin{tcblisting}{listing only}
You'll be given a set of sentences to read through carefully. Once you've reviewed them, I'll
ask you a question related to the information in those sentences. Your job is to think
critically about the details, analyze the sentences in relation to the question, and then
provide your answer. If the information clearly supports a partial answer, provide that.
However, if the evidence is unclear or insufficient, it is okay to respond with "No answer."

**Input:**
- **Sentences:**
```
{context}
```

- **Question:**
```
{question}
```

**Response:**
- **Reasoning:**
  - [Describe how you thought through the sentences and how they helped you reach your conclusion. If the evidence is unclear or insufficient to provide a reliable answer, explain why. Your reasoning should not exceed 10,000 words.]
- **Answer:** [Provide an answer only if it is clearly supported by the information in the sentences. If the evidence is unclear or insufficient, respond with "No answer."]
\end{tcblisting}

\subsubsection{Prompt template for LLM-as-a-Judge}
\begin{tcblisting}{listing only}
You will be given a question along with a response generated by an assistant and the
corresponding ground truth data. Your task is to assess the response based on its accuracy
and completeness in comparison to the ground truth. For each entry in the ground truth,
determine whether the information provided by the assistant is an "Exact Match," a "Partial
Match," or a "No Match."

#### **Evaluation Criteria:**

- **Exact Match**: The assistant's response precisely matches the ground truth in both content and detail.
- **Partial Match**: The assistant's response includes some correct information but is either incomplete, incorrectly ordered, or contains inaccuracies.
- **No Match**: The assistant's response does not accurately reflect the ground truth or is missing entirely.

#### **Special Cases:**

**Ground Truth is None**:  
- If the ground truth is `None` (represented as an empty list `[]`):
  - **Exact Match**: If the assistant's response indicates that there is no information or content.
  - **No Match**: If the assistant's response provides any information when the ground truth is `None`.

#### **Output Format:**

- The output should be a list of objects where each object contains:
  - An `"id"` that matches the `id` of the corresponding ground truth entry.
  - A `"label"` indicating whether the assistant's response is an `"Exact Match"`, `"Partial Match"`, or `"No Match"`.

- The number of output objects should match the number of entries in the ground truth.

---

### **Examples:**
{in_context_examples}

====== Your task starts here ======

**Question:**
```
{question}
```

**Assistant's Response:**
```
{pred}
```

**Ground Truth:**
```
{gold}
```

**Output Format:**
```
{output_format}
```
\end{tcblisting}

\newpage
\subsection{Detailed experimental setup}
\label{sec:appendix-experimental}
We set temperature and top-p parameters to $0.0$ and $1.0$, respectively, for all our experiments. For Gemini-2.5-Flash, we set its thinking budget to $0$. 
Table \ref{tab:models} lists the models used in our experiments.

\begin{table*}[h]
  \centering
  \resizebox{\linewidth}{!}{
  \begin{tabular}{lrrlc}
      \toprule
      \textbf{Model} & \textbf{Size} & \textbf{Context} & \textbf{Model name / snapshot} & \textbf{License} \\
      \midrule
      GPT-4o & --- & 128k & \texttt{gpt-4o-2024-08-06} & OpenAI Service Terms\footnotemark[3] \\
      GPT-4o-mini & --- & 128k & \texttt{gpt-4o-mini-2024-07-18} & OpenAI Service Terms\\
      Gemini-2.5-Flash & --- & 1M & \texttt{gemini-2.5-flash-preview-04-17} & Gemini API Additional Terms of Service\footnotemark[4]\\
      Llama-4-Maverick & 400B & 1M & \texttt{meta-llama/Llama-4-Maverick-17B-128E-Instruct} & Llama 4 Community License Agreement\footnotemark[5] \\ 
      Llama-4-Scout & 109B & 10M & \texttt{meta-llama/Llama-4-Scout-17B-16E-Instruct} & Llama 4 Community License Agreement \\
      \bottomrule
  \end{tabular}
  }
  \caption{A list of the LLMs used in the experiments.
  An em-dash (---) means that the model size is not publicly disclosed.
  }
  \label{tab:models}
\end{table*}

\footnotetext[3]{\url{https://openai.com/policies/services-agreement/} [Accessed: May 12, 2025]}
\footnotetext[4]{\url{https://ai.google.dev/gemini-api/terms} [Accessed: May 12, 2025] }
\footnotetext[5]{\url{https://www.llama.com/llama4/license/} [Accessed: May 12, 2025]}

\clearpage
\subsection{Full results}
\label{sec:appendix-full-results}
\subsubsection{Factoid QA tasks (BGE embeddings)}
\input{results/qa-gpt-4o-mini}
\input{results/qa-gpt-4o}
\input{results/qa-gemini-2.5-flash}
\input{results/qa-llama4-scout}
\input{results/qa-llama4-maverick}

\clearpage
\subsubsection{HoloBench (BGE embeddings)}
\input{results/holobench-gpt-4o-mini}
\input{results/holobench-gpt-4o}
\input{results/holobench-gemini}
\input{results/holobench-llama4-scout}
\input{results/holobench-llama4-maverick}

\clearpage
\subsubsection{Factoid QA tasks (GTE embeddings)}
\label{app:sssec:factoid_gte}
\input{results/qa-gpt-4o-gte}

\clearpage
\subsubsection{HoloBench (GTE embeddings)}
\label{appLsssec:holobench_gte}
\input{results/holobench-gpt-4o-gte}

%% file: results/qa-gpt-4o-mini.tex
\begin{table*}[h]
\centering
\setlength{\tabcolsep}{6pt}
\small
\begin{tabular}{
l  
l  
r  
r  
r  
r  
r  
}
\toprule
Task & Method & Score & Context recall & Reduction (\%) & $n_\text{in}$ & $n_\text{out}$ \\
\midrule
\multirow{9}{4em}{HotpotQA} & zeroshot & 39 & 0.00 ± 0.00 & 100.00 ± 0.00 & 63.03 ± 6.98 & 15.56 ± 9.92 \\
& fixed-1k & 60 & 69.33 ± 30.77 & 99.33 ± 0.09 & 852.30 ± 57.33 & 20.53 ± 13.61 \\
& fixed-5k & 66 & 84.50 ± 26.93 & 96.46 ± 0.29 & 3983.81 ± 219.82 & 20.87 ± 14.45 \\
& fixed-10k & 66 & 88.50 ± 23.05 & 92.89 ± 0.55 & 7911.32 ± 440.28 & 21.85 ± 14.35 \\
& fixed-25k & 67 & 92.50 ± 20.15 & 82.19 ± 1.32 & 19763.59 ± 1109.32 & 22.36 ± 14.48 \\
& fixed-50k & 66 & 95.33 ± 14.23 & 64.46 ± 2.45 & 39597.66 ± 2224.65 & 24.00 ± 15.87 \\
& full-context & 45 & 100.00 ± 0.00 & 0.00 ± 0.00 & 109666.92 ± 5537.59 & 16.49 ± 18.35 \\
 \cmidrule{2-7}& \textsc{self-route} & 61 & 90.83 ± 22.77 & 75.25 ± 40.17 & 28008.57 ± 45573.55 & 17.31 ± 17.84 \\
& adaptive-\textit{k} & \textbf{63} & 70.83 ± 31.01 & 99.24 ± 0.17 & 954.33 ± 206.78 & 20.46 ± 14.43 \\
 \midrule\multirow{9}{4em}{NQ} & zeroshot & 49 & 0.00 ± 0.00 & 100.00 ± 0.00 & 53.37 ± 2.29 & 23.36 ± 15.22 \\
& fixed-1k & 54 & 26.45 ± 30.50 & 99.36 ± 0.09 & 806.77 ± 70.12 & 28.50 ± 18.53 \\
& fixed-5k & 59 & 42.45 ± 36.68 & 96.66 ± 0.28 & 3837.48 ± 333.10 & 31.33 ± 21.80 \\
& fixed-10k & 58 & 50.35 ± 35.94 & 93.27 ± 0.52 & 7632.00 ± 655.62 & 32.11 ± 23.94 \\
& fixed-25k & 62 & 62.78 ± 32.87 & 83.10 ± 1.25 & 19051.28 ± 1574.22 & 33.91 ± 26.90 \\
& fixed-50k & 59 & 68.89 ± 29.96 & 66.16 ± 2.45 & 38102.93 ± 2989.91 & 37.39 ± 29.33 \\
& full-context & 41 & 100.00 ± 0.00 & 0.00 ± 0.00 & 110607.54 ± 4711.16 & 23.10 ± 27.28 \\
 \cmidrule{2-7}& \textsc{self-route} & \textbf{55} & 52.72 ± 36.37 & 77.33 ± 38.86 & 25839.53 ± 44249.67 & 23.60 ± 20.09 \\
& adaptive-\textit{k} & 54 & 27.20 ± 31.58 & 99.25 ± 0.25 & 927.69 ± 283.52 & 28.83 ± 18.99 \\
 \midrule\multirow{9}{4em}{TriviaQA} & zeroshot & 83 & 0.00 ± 0.00 & 100.00 ± 0.00 & 60.09 ± 7.79 & 7.85 ± 6.27 \\
& fixed-1k & 94 & 31.21 ± 36.58 & 99.34 ± 0.10 & 833.75 ± 67.42 & 11.86 ± 9.07 \\
& fixed-5k & 93 & 42.10 ± 39.93 & 96.53 ± 0.26 & 3913.01 ± 260.44 & 11.73 ± 9.41 \\
& fixed-10k & 92 & 49.90 ± 40.11 & 93.03 ± 0.49 & 7772.18 ± 488.60 & 12.64 ± 11.19 \\
& fixed-25k & 93 & 54.74 ± 40.26 & 82.59 ± 1.14 & 19384.51 ± 1164.94 & 13.64 ± 11.10 \\
& fixed-50k & 94 & 61.66 ± 37.83 & 65.21 ± 2.27 & 38819.58 ± 2326.33 & 15.72 ± 11.95 \\
& full-context & 61 & 100.00 ± 0.00 & 0.00 ± 0.00 & 110733.69 ± 3419.97 & 11.95 ± 13.89 \\
 \cmidrule{2-7}& \textsc{self-route} & 90 & 48.19 ± 39.84 & 84.95 ± 31.53 & 17112.00 ± 35900.97 & 8.69 ± 8.41 \\
& adaptive-\textit{k} & \textbf{92} & 31.21 ± 36.58 & 99.26 ± 0.23 & 918.86 ± 240.86 & 11.69 ± 9.15 \\
 \midrule
\multirow{9}{4em}{Average} & zeroshot & 57.00 & 0.00 & 0.00 & 58.83 & 15.59 \\
& fixed-1k & 69.33 & 42.33 & 99.34 & 830.94 & 20.30 \\
& fixed-5k & 72.67 & 56.35 & 96.55 & 3911.43 & 21.31 \\
& fixed-10k & 72.00 & 62.92 & 93.07 & 7771.83 & 22.20 \\
& fixed-25k & 74.00 & 70.00 & 82.63 & 19399.79 & 23.30 \\
& fixed-50k & 73.00 & 75.29 & 65.28 & 38840.06 & 25.70 \\
& full-context & 49.00 & 100.00 & 0.00 & 110336.05 & 17.18 \\ \cmidrule{2-7}
& \textsc{self-route} & 68.67 & 63.91 & 79.18 & 23653.37 & 16.53 \\
& adaptive-\textit{k} & \textbf{69.67} & 43.08 & 99.25 & 933.63 & 20.33 \\
\bottomrule
\end{tabular}
\caption{Full GPT-4o-mini's results in the factoid QA tasks.}
\label{tab:qa-gpt-4o-mini}
\end{table*}

%% file: results/qa-gpt-4o.tex
\begin{table*}[h]
\centering
\setlength{\tabcolsep}{6pt}
\small
\begin{tabular}{
l  
l  
r  
r  
r  
r  
r  
}
\toprule
Task & Method & Score & Context recall & Reduction (\%) & $n_\text{in}$ & $n_\text{out}$ \\
\midrule
\multirow{9}{4em}{HotpotQA} & zeroshot & 50 & 0.00 ± 0.00 & 100.00 ± 0.00 & 63.03 ± 6.98 & 19.31 ± 16.53 \\
& fixed-1k & 61 & 69.33 ± 30.77 & 99.33 ± 0.09 & 852.30 ± 57.33 & 27.72 ± 18.35 \\
& fixed-5k & 70 & 84.50 ± 26.93 & 96.46 ± 0.29 & 3983.81 ± 219.82 & 27.25 ± 17.25 \\
& fixed-10k & 76 & 88.50 ± 23.05 & 92.89 ± 0.55 & 7911.32 ± 440.28 & 29.16 ± 18.31 \\
& fixed-25k & 74 & 92.50 ± 20.15 & 82.19 ± 1.32 & 19763.59 ± 1109.32 & 28.66 ± 20.86 \\
& fixed-50k & 73 & 95.33 ± 14.23 & 64.46 ± 2.45 & 39597.66 ± 2224.65 & 27.89 ± 20.22 \\
& full-context & 48 & 100.00 ± 0.00 & 0.00 ± 0.00 & 109666.92 ± 5537.59 & 18.91 ± 20.30 \\
 \cmidrule{2-7}& \textsc{self-route} & \textbf{66} & 84.50 ± 26.93 & 96.46 ± 0.29 & 23663.11 ± 42241.38 & 22.49 ± 20.04 \\
& adaptive-\textit{k} & 63 & 70.83 ± 31.01 & 99.24 ± 0.17 & 954.33 ± 206.78 & 28.24 ± 19.56 \\
 \midrule\multirow{9}{4em}{NQ} & zeroshot & 57 & 0.00 ± 0.00 & 100.00 ± 0.00 & 53.37 ± 2.29 & 27.38 ± 24.23 \\
& fixed-1k & 59 & 26.45 ± 30.50 & 99.36 ± 0.09 & 806.77 ± 70.12 & 33.62 ± 26.13 \\
& fixed-5k & 64 & 42.45 ± 36.68 & 96.66 ± 0.28 & 3837.48 ± 333.10 & 37.37 ± 30.24 \\
& fixed-10k & 64 & 50.35 ± 35.94 & 93.27 ± 0.52 & 7632.00 ± 655.62 & 38.41 ± 32.61 \\
& fixed-25k & 64 & 62.78 ± 32.87 & 83.10 ± 1.25 & 19051.28 ± 1574.22 & 38.48 ± 33.35 \\
& fixed-50k & 63 & 68.89 ± 29.96 & 66.16 ± 2.45 & 38102.93 ± 2989.91 & 39.76 ± 33.99 \\
& full-context & 41 & 100.00 ± 0.00 & 0.00 ± 0.00 & 110607.54 ± 4711.16 & 24.74 ± 33.79 \\
 \cmidrule{2-7}& \textsc{self-route} & \textbf{61} & 42.45 ± 36.68 & 96.66 ± 0.28 & 24713.66 ± 43308.30 & 33.64 ± 34.15 \\
& adaptive-\textit{k} & \textbf{61} & 27.20 ± 31.58 & 99.25 ± 0.25 & 927.69 ± 283.52 & 34.85 ± 27.40 \\
 \midrule\multirow{9}{4em}{TriviaQA} & zeroshot & 91 & 0.00 ± 0.00 & 100.00 ± 0.00 & 60.09 ± 7.79 & 8.33 ± 7.16 \\
& fixed-1k & 96 & 31.21 ± 36.58 & 99.34 ± 0.10 & 833.75 ± 67.42 & 16.10 ± 11.89 \\
& fixed-5k & 95 & 42.10 ± 39.93 & 96.53 ± 0.26 & 3913.01 ± 260.44 & 16.01 ± 11.49 \\
& fixed-10k & 94 & 49.90 ± 40.11 & 93.03 ± 0.49 & 7772.18 ± 488.60 & 15.55 ± 9.91 \\
& fixed-25k & 93 & 54.74 ± 40.26 & 82.59 ± 1.14 & 19384.51 ± 1164.94 & 15.96 ± 9.81 \\
& fixed-50k & 93 & 61.66 ± 37.83 & 65.21 ± 2.27 & 38819.58 ± 2326.33 & 16.31 ± 10.62 \\
& full-context & 62 & 100.00 ± 0.00 & 0.00 ± 0.00 & 110733.69 ± 3419.97 & 12.94 ± 12.48 \\
 \cmidrule{2-7}& \textsc{self-route} & 92 & 42.10 ± 39.93 & 96.53 ± 0.26 & 13900.19 ± 31870.44 & 11.39 ± 10.31 \\
& adaptive-\textit{k} & \textbf{96} & 31.21 ± 36.58 & 99.26 ± 0.23 & 918.86 ± 240.86 & 16.10 ± 12.07 \\
 \midrule
\multirow{9}{4em}{Average} & zeroshot & 66.00 & 0.00 & 0.00 & 58.83 & 18.34 \\
& fixed-1k & 72.00 & 42.33 & 99.34 & 830.94 & 25.81 \\
& fixed-5k & 76.33 & 56.35 & 96.55 & 3911.43 & 26.88 \\
& fixed-10k & 78.00 & 62.92 & 93.07 & 7771.83 & 27.71 \\
& fixed-25k & 77.00 & 70.00 & 82.63 & 19399.79 & 27.70 \\
& fixed-50k & 76.33 & 75.29 & 65.28 & 38840.06 & 27.99 \\
& full-context & 50.33 & 100.00 & 0.00 & 110336.05 & 18.86 \\ \cmidrule{2-7}
& \textsc{self-route} & 73.00 & 56.35 & 96.55 & 20758.99 & 22.51 \\
& adaptive-\textit{k} & \textbf{73.33} & 43.08 & 99.25 & 933.63 & 26.40 \\
\bottomrule
\end{tabular}
\caption{Full GPT-4o's results in the factoid QA tasks.}
\label{tab:qa-gpt-4o}
\end{table*}

%% file: results/qa-gemini-2.5-flash.tex
\begin{table*}[h]
\centering
\setlength{\tabcolsep}{6pt}
\small
\begin{tabular}{
l  
l  
r  
r  
r  
r  
r  
}
\toprule
Task & Method & Score & Context recall & Reduction (\%) & $n_\text{in}$ & $n_\text{out}$ \\
\midrule
\multirow{9}{4em}{HotpotQA} & zeroshot & 46 & 0.00 ± 0.00 & 100.00 ± 0.00 & 57.25 ± 7.72 & 3.00 ± 1.62 \\
& fixed-1k & 54 & 69.33 ± 30.77 & 99.33 ± 0.09 & 883.55 ± 67.88 & 15.24 ± 23.19 \\
& fixed-5k & 63 & 84.50 ± 26.93 & 96.46 ± 0.29 & 4170.89 ± 274.70 & 12.92 ± 19.22 \\
& fixed-10k & 66 & 88.50 ± 23.05 & 92.89 ± 0.55 & 8295.62 ± 548.91 & 12.62 ± 17.17 \\
& fixed-25k & 66 & 92.50 ± 20.15 & 82.19 ± 1.32 & 20727.06 ± 1379.61 & 15.09 ± 20.48 \\
& fixed-50k & 72 & 95.33 ± 14.23 & 64.46 ± 2.45 & 41530.89 ± 2780.34 & 16.97 ± 22.12 \\
& full-context & 71 & 100.00 ± 0.00 & 0.00 ± 0.00 & 115121.31 ± 5964.79 & 17.60 ± 19.32 \\
 \cmidrule{2-7}& \textsc{self-route} & \textbf{68} & 95.33 ± 17.42 & 69.45 ± 43.53 & 36820.47 ± 52663.73 & 13.95 ± 19.00 \\
& adaptive-\textit{k} & 55 & 70.83 ± 31.01 & 99.24 ± 0.17 & 990.43 ± 216.72 & 15.29 ± 24.47 \\
 \midrule\multirow{9}{4em}{NQ} & zeroshot & 47 & 0.00 ± 0.00 & 100.00 ± 0.00 & 46.30 ± 2.46 & 4.67 ± 3.92 \\
& fixed-1k & 44 & 26.45 ± 30.50 & 99.36 ± 0.09 & 826.53 ± 80.39 & 26.14 ± 33.25 \\
& fixed-5k & 59 & 42.45 ± 36.68 & 96.66 ± 0.28 & 3959.77 ± 351.93 & 31.82 ± 32.77 \\
& fixed-10k & 59 & 50.35 ± 35.94 & 93.27 ± 0.52 & 7891.38 ± 704.11 & 115.35 ± 816.66 \\
& fixed-25k & 62 & 62.78 ± 32.87 & 83.10 ± 1.25 & 19730.89 ± 1697.34 & 34.87 ± 53.13 \\
& fixed-50k & 61 & 68.89 ± 29.96 & 66.16 ± 2.45 & 39505.22 ± 3266.78 & 35.23 ± 47.76 \\
& full-context & 64 & 100.00 ± 0.00 & 0.00 ± 0.00 & 115142.91 ± 5115.01 & 28.15 ± 22.01 \\
 \cmidrule{2-7}& \textsc{self-route} & \textbf{60} & 54.71 ± 36.63 & 74.41 ± 40.87	 & 30547.34 ± 48902.74 & 27.40 ± 30.96 \\
& adaptive-\textit{k} & 47 & 27.20 ± 31.58 & 99.25 ± 0.25 & 951.38 ± 295.47 & 29.04 ± 32.84 \\
 \midrule\multirow{9}{4em}{TriviaQA} & zeroshot & 93 & 0.00 ± 0.00 & 100.00 ± 0.00 & 54.36 ± 8.36 & 2.55 ± 1.48 \\
& fixed-1k & 87 & 31.21 ± 36.58 & 99.34 ± 0.10 & 859.44 ± 78.27 & 8.99 ± 16.82 \\
& fixed-5k & 93 & 42.10 ± 39.93 & 96.53 ± 0.26 & 4073.06 ± 317.10 & 8.74 ± 15.25 \\
& fixed-10k & 92 & 49.90 ± 40.11 & 93.03 ± 0.49 & 8097.22 ± 589.74 & 9.30 ± 14.94 \\
& fixed-25k & 93 & 54.74 ± 40.26 & 82.59 ± 1.14 & 20213.64 ± 1424.47 & 8.41 ± 11.68 \\
& fixed-50k & 92 & 61.66 ± 37.83 & 65.21 ± 2.27 & 40490.49 ± 2852.03 & 11.95 ± 14.79 \\
& full-context & 95 & 100.00 ± 0.00 & 0.00 ± 0.00 & 115669.41 ± 4229.98 & 10.48 ± 8.73 \\
 \cmidrule{2-7}& \textsc{self-route} & \textbf{94} & 49.19 ± 39.24 & 84.95 ± 31.53 & 17948.59 ± 37786.95 & 7.39 ± 10.84 \\
& adaptive-\textit{k} & 86 & 31.21 ± 36.58 & 99.26 ± 0.23 & 947.50 ± 244.93 & 7.53 ± 13.74 \\
 \midrule
\multirow{9}{4em}{Average} & zeroshot & 62.00 & 0.00 & 0.00 & 52.64 & 3.41 \\
& fixed-1k & 61.67 & 42.33 & 99.34 & 856.51 & 16.79 \\
& fixed-5k & 71.67 & 56.35 & 96.55 & 4067.91 & 17.83 \\
& fixed-10k & 72.33 & 62.92 & 93.07 & 8094.74 & 45.76 \\
& fixed-25k & 73.67 & 70.00 & 82.63 & 20223.86 & 19.46 \\
& fixed-50k & 75.00 & 75.29 & 65.28 & 40508.87 & 21.38 \\
& full-context & 76.67 & 100.00 & 0.00 & 115311.21 & 18.74 \\ \cmidrule{2-7}
& \textsc{self-route} & \textbf{74.00} & 66.41 & 76.27 & 28438.80 & 16.25 \\
& adaptive-\textit{k} & 62.67 & 43.08 & 99.25 & 963.10 & 17.29 \\
\bottomrule
\end{tabular}
\caption{Full Gemini-2.5-Flash's results in the factoid QA tasks.}
\label{tab:qa-gemini}
\end{table*}

%% file: results/qa-llama4-scout.tex
\begin{table*}[h]
\centering
\setlength{\tabcolsep}{6pt}
\small
\begin{tabular}{
l  
l  
r  
r  
r  
r  
r  
}
\toprule
Task & Method & Score & Context recall & Reduction (\%) & $n_\text{in}$ & $n_\text{out}$ \\
\midrule
\multirow{9}{4em}{HotpotQA} & zeroshot & 38 & 0.00 ± 0.00 & 100.00 ± 0.00 & 61.17 ± 7.05 & 226.87 ± 1635.13 \\
& fixed-1k & 65 & 69.33 ± 30.77 & 99.33 ± 0.09 & 850.33 ± 59.50 & 59.78 ± 97.50 \\
& fixed-5k & 65 & 84.50 ± 26.93 & 96.46 ± 0.29 & 4006.32 ± 227.71 & 41.23 ± 63.66 \\
& fixed-10k & 68 & 88.50 ± 23.05 & 92.89 ± 0.55 & 7963.66 ± 456.87 & 38.49 ± 60.86 \\
& fixed-25k & 68 & 92.50 ± 20.15 & 82.19 ± 1.32 & 19912.13 ± 1152.36 & 33.14 ± 50.95 \\
& fixed-50k & 67 & 95.33 ± 14.23 & 64.46 ± 2.45 & 39898.83 ± 2310.69 & 36.25 ± 56.47 \\
& full-context & 67 & 100.00 ± 0.00 & 0.00 ± 0.00 & 110457.05 ± 5390.30 & 36.77 ± 65.52 \\
 \cmidrule{2-7}& \textsc{self-route} & \textbf{73} & 89.50 ± 23.53 & 84.89 ± 31.51 & 17292.33 ± 36140.60 & 60.39 ± 78.22 \\
& adaptive-\textit{k} & 63 & 70.83 ± 31.01 & 99.24 ± 0.17 & 952.64 ± 206.19 & 53.92 ± 87.49 \\
 \midrule\multirow{9}{4em}{NQ} & zeroshot & 58 & 0.00 ± 0.00 & 100.00 ± 0.00 & 51.54 ± 2.41 & 62.84 ± 53.57 \\
& fixed-1k & 62 & 26.45 ± 30.50 & 99.36 ± 0.09 & 802.11 ± 71.62 & 60.35 ± 48.38 \\
& fixed-5k & 66 & 42.45 ± 36.68 & 96.66 ± 0.28 & 3847.40 ± 341.79 & 73.75 ± 67.45 \\
& fixed-10k & 64 & 50.35 ± 35.94 & 93.27 ± 0.52 & 7660.11 ± 675.79 & 76.13 ± 61.56 \\
& fixed-25k & 66 & 62.78 ± 32.87 & 83.10 ± 1.25 & 19136.92 ± 1631.03 & 85.91 ± 81.62 \\
& fixed-50k & 68 & 68.89 ± 29.96 & 66.16 ± 2.45 & 38284.28 ± 3103.43 & 288.08 ± 1664.19 \\
& full-context & 68 & 100.00 ± 0.00 & 0.00 ± 0.00 & 111154.46 ± 4416.57 & 113.88 ± 202.07 \\
 \cmidrule{2-7}& \textsc{self-route} & \textbf{66} & 48.21 ± 36.70 & 85.06 ± 31.57 & 17228.44 ± 36372.48 & 142.10 ± 316.76 \\
& adaptive-\textit{k} & 61 & 27.20 ± 31.58 & 99.25 ± 0.25 & 923.58 ± 286.51 & 67.89 ± 59.55 \\
 \midrule\multirow{9}{4em}{TriviaQA} & zeroshot & 85 & 0.00 ± 0.00 & 100.00 ± 0.00 & 58.29 ± 7.89 & 20.12 ± 39.32 \\
& fixed-1k & 98 & 31.21 ± 36.58 & 99.34 ± 0.10 & 830.54 ± 66.35 & 24.97 ± 54.24 \\
& fixed-5k & 98 & 42.10 ± 39.93 & 96.53 ± 0.26 & 3928.14 ± 261.91 & 24.80 ± 63.48 \\
& fixed-10k & 97 & 49.90 ± 40.11 & 93.03 ± 0.49 & 7808.89 ± 495.48 & 18.29 ± 42.28 \\
& fixed-25k & 96 & 54.74 ± 40.26 & 82.59 ± 1.14 & 19486.58 ± 1189.90 & 19.47 ± 51.48 \\
& fixed-50k & 95 & 61.66 ± 37.83 & 65.21 ± 2.27 & 39032.69 ± 2399.86 & 12.60 ± 22.71 \\
& full-context & 96 & 100.00 ± 0.00 & 0.00 ± 0.00 & 111322.11 ± 3373.09 & 23.69 ± 52.77 \\
 \cmidrule{2-7}& \textsc{self-route} & 97 & 47.69 ± 40.13 & 89.78 ± 24.76 & 11742.38 ± 28544.35 & 34.38 ± 63.98 \\
& adaptive-\textit{k} & \textbf{98} & 31.21 ± 36.58 & 99.26 ± 0.23 & 915.89 ± 238.98 & 30.32 ± 73.20 \\
 \midrule
\multirow{9}{4em}{Average} & zeroshot & 60.33 & 0.00 & 0.00 & 57.00 & 103.28 \\
& fixed-1k & 75.00 & 42.33 & 99.34 & 827.66 & 48.37 \\
& fixed-5k & 76.33 & 56.35 & 96.55 & 3927.29 & 46.59 \\
& fixed-10k & 76.33 & 62.92 & 93.07 & 7810.89 & 44.30 \\
& fixed-25k & 76.67 & 70.00 & 82.63 & 19511.88 & 46.17 \\
& fixed-50k & 76.67 & 75.29 & 65.28 & 39071.93 & 112.31 \\
& full-context & 77.00 & 100.00 & 0.00 & 110977.87 & 58.11 \\ \cmidrule{2-7}
& \textsc{self-route} & \textbf{78.67} & 61.80 & 86.57 & 15421.05 & 78.96 \\
& adaptive-\textit{k} & 74.00 & 43.08 & 99.25 & 930.70 & 50.71 \\
\bottomrule
\end{tabular}
\caption{Full Llama4-Scout's results in the factoid QA tasks.}
\label{tab:qa-llama4-scout}
\end{table*}

%% file: results/qa-llama4-maverick.tex
\begin{table*}[h]
\centering
\setlength{\tabcolsep}{6pt}
\small
\begin{tabular}{
l  
l  
r  
r  
r  
r  
r  
}
\toprule
Task & Method & Score & Context recall & Reduction (\%) & $n_\text{in}$ & $n_\text{out}$ \\
\midrule
\multirow{9}{4em}{HotpotQA} & zeroshot & 52 & 0.00 ± 0.00 & 100.00 ± 0.00 & 61.17 ± 7.05 & 99.43 ± 110.61 \\
& fixed-1k & 71 & 69.33 ± 30.77 & 99.33 ± 0.09 & 850.33 ± 59.50 & 110.04 ± 141.75 \\
& fixed-5k & 78 & 84.50 ± 26.93 & 96.46 ± 0.29 & 4006.32 ± 227.71 & 68.99 ± 88.75 \\
& fixed-10k & 74 & 88.50 ± 23.05 & 92.89 ± 0.55 & 7963.66 ± 456.87 & 44.12 ± 63.11 \\
& fixed-25k & 71 & 92.50 ± 20.15 & 82.19 ± 1.32 & 19912.13 ± 1152.36 & 43.30 ± 55.48 \\
& fixed-50k & 72 & 95.33 ± 14.23 & 64.46 ± 2.45 & 39898.83 ± 2310.69 & 45.51 ± 66.82 \\
& full-context & 75 & 100.00 ± 0.00 & 0.00 ± 0.00 & 110457.05 ± 5390.30 & 41.18 ± 54.88 \\
 \cmidrule{2-7}& \textsc{self-route} & \textbf{79} & 86.00 ± 26.35 & 92.61 ± 19.00 & 8383.42 ± 21450.47 & 90.55 ± 120.15 \\
& adaptive-\textit{k} & 71 & 70.83 ± 31.01 & 99.24 ± 0.17 & 952.64 ± 206.19 & 112.62 ± 146.43 \\
 \midrule\multirow{9}{4em}{NQ} & zeroshot & 53 & 0.00 ± 0.00 & 100.00 ± 0.00 & 51.54 ± 2.41 & 54.22 ± 54.55 \\
& fixed-1k & 63 & 26.45 ± 30.50 & 99.36 ± 0.09 & 802.11 ± 71.62 & 68.16 ± 58.41 \\
& fixed-5k & 65 & 42.45 ± 36.68 & 96.66 ± 0.28 & 3847.40 ± 341.79 & 75.36 ± 69.75 \\
& fixed-10k & 67 & 50.35 ± 35.94 & 93.27 ± 0.52 & 7660.11 ± 675.79 & 75.90 ± 79.63 \\
& fixed-25k & 64 & 62.78 ± 32.87 & 83.10 ± 1.25 & 19136.92 ± 1631.03 & 69.41 ± 69.85 \\
& fixed-50k & 64 & 68.89 ± 29.96 & 66.16 ± 2.45 & 38284.28 ± 3103.43 & 71.23 ± 77.03 \\
& full-context & 67 & 100.00 ± 0.00 & 0.00 ± 0.00 & 111154.46 ± 4416.57 & 66.25 ± 66.57 \\
 \cmidrule{2-7}& \textsc{self-route} & \textbf{65} & 45.95 ± 37.23 & 90.86 ± 23.07 & 10536.57 ± 26530.78 & 69.87 ± 75.65 \\
& adaptive-\textit{k} & 62 & 27.20 ± 31.58 & 99.25 ± 0.25 & 923.58 ± 286.51 & 75.56 ± 73.03 \\
 \midrule\multirow{9}{4em}{TriviaQA} & zeroshot & 91 & 0.00 ± 0.00 & 100.00 ± 0.00 & 58.29 ± 7.89 & 25.52 ± 54.17 \\
& fixed-1k & 96 & 31.21 ± 36.58 & 99.34 ± 0.10 & 830.54 ± 66.35 & 39.52 ± 57.18 \\
& fixed-5k & 98 & 42.10 ± 39.93 & 96.53 ± 0.26 & 3928.14 ± 261.91 & 31.02 ± 48.63 \\
& fixed-10k & 98 & 49.90 ± 40.11 & 93.03 ± 0.49 & 7808.89 ± 495.48 & 21.72 ± 32.93 \\
& fixed-25k & 96 & 54.74 ± 40.26 & 82.59 ± 1.14 & 19486.58 ± 1189.90 & 22.79 ± 46.88 \\
& fixed-50k & 96 & 61.66 ± 37.83 & 65.21 ± 2.27 & 39032.69 ± 2399.86 & 15.46 ± 34.71 \\
& full-context & 96 & 100.00 ± 0.00 & 0.00 ± 0.00 & 111322.11 ± 3373.09 & 19.58 ± 34.15 \\
 \cmidrule{2-7}& \textsc{self-route} & \textbf{98} & 44.60 ± 40.03 & 93.64 ± 16.55 & 7440.28 ± 19973.57 & 37.87 ± 63.00 \\
& adaptive-\textit{k} & 96 & 31.21 ± 36.58 & 99.26 ± 0.23 & 915.89 ± 238.98 & 49.73 ± 70.11 \\
 \midrule
\multirow{9}{4em}{Average} & zeroshot & 65.33 & 0.00 & 0.00 & 57.00 & 59.72 \\
& fixed-1k & 76.67 & 42.33 & 99.34 & 827.66 & 72.57 \\
& fixed-5k & 80.33 & 56.35 & 96.55 & 3927.29 & 58.46 \\
& fixed-10k & 79.67 & 62.92 & 93.07 & 7810.89 & 47.25 \\
& fixed-25k & 77.00 & 70.00 & 82.63 & 19511.88 & 45.17 \\
& fixed-50k & 77.33 & 75.29 & 65.28 & 39071.93 & 44.07 \\
& full-context & 79.33 & 100.00 & 0.00 & 110977.87 & 42.34 \\ \cmidrule{2-7}
& \textsc{self-route} & \textbf{80.67} & 58.85 & 92.37 & 8786.76 & 66.10 \\
& adaptive-\textit{k} & 76.33 & 43.08 & 99.25 & 930.70 & 79.30 \\
\bottomrule
\end{tabular}
\caption{Full Llama4-Maverick's results in the factoid QA tasks.}
\label{tab:qa-llama4-maverick}
\end{table*}

%% file: results/holobench-gpt-4o-mini.tex
\begin{table*}[h]
\centering
\setlength{\tabcolsep}{6pt}
\small
\begin{tabular}{
l  
l  
r  
r  
r  
r  
r  
}
\toprule
Info amount & Method & Score & Context recall & Reduction (\%) & $n_\text{in}$ & $n_\text{out}$ \\
\midrule
\multirow{9}{4em}{info5k} & zeroshot & 10.00 & 0.00 ± 0.00 & 100.00 ± 0.00 & 58.13 & 43.34 \\
& fixed-1k & 28.19 & 12.05 ± 6.42 & 99.18 ± 0.43 & 1000.07 & 325.14 \\
& fixed-5k & 38.74 & 51.92 ± 29.82 & 95.85 ± 1.86 & 4194.01 & 923.67 \\
& fixed-10k & 43.50 & 66.68 ± 30.73 & 91.80 ± 2.92 & 8011.64 & 801.81 \\
& fixed-25k & 39.81 & 78.48 ± 26.96 & 79.59 ± 4.56 & 19574.94 & 1489.67 \\
& fixed-50k & 37.67 & 86.79 ± 20.88 & 57.76 ± 5.92 & 39224.80 & 2342.40 \\
& full-context & 37.76 & 100.00 ± 0.00 & 0.00 ± 0.00 & 85882.50 & 3147.27 \\
 \cmidrule{2-7}& \textsc{self-route} & 31.32 & 69.46 ± 27.62 & 74.68 ± 40.18 & 23044.97 & 1523.99 \\
& adaptive-\textit{k} & \textbf{40.86} & 75.74 ± 30.48 & 74.07 ± 25.68 & 24625.02 & 2220.94 \\
 \midrule\multirow{9}{4em}{info10k} & zeroshot & 6.22 & 0.00 ± 0.00 & 100.00 ± 0.00 & 58.13 & 45.28 \\
& fixed-1k & 22.55 & 6.53 ± 3.11 & 99.19 ± 0.40 & 1003.83 & 322.87 \\
& fixed-5k & 34.06 & 31.77 ± 14.86 & 95.84 ± 1.90 & 4228.71 & 987.09 \\
& fixed-10k & 34.85 & 59.10 ± 27.14 & 91.74 ± 3.47 & 8235.02 & 1712.06 \\
& fixed-25k & 36.44 & 78.18 ± 26.55 & 79.55 ± 5.42 & 19838.92 & 1716.79 \\
& fixed-50k & 29.98 & 87.34 ± 20.51 & 57.88 ± 6.41 & 39437.42 & 2910.81 \\
& full-context & 26.59 & 100.00 ± 0.00 & 0.00 ± 0.00 & 86139.74 & 4370.90 \\
 \cmidrule{2-7}& \textsc{self-route} & 28.56 & 48.18 ± 27.16 & 77.76 ± 37.78 & 19627.93 & 2228.59 \\
& adaptive-\textit{k} & \textbf{33.16} & 68.54 ± 32.55 & 79.22 ± 21.59 & 20233.99 & 2338.73 \\
 \midrule\multirow{9}{4em}{info25k} & zeroshot & 4.22 & 0.00 ± 0.00 & 100.00 ± 0.00 & 58.13 & 43.17 \\
& fixed-1k & 16.67 & 2.77 ± 1.15 & 99.21 ± 0.32 & 999.62 & 331.77 \\
& fixed-5k & 25.76 & 14.06 ± 5.50 & 95.96 ± 1.56 & 4215.72 & 670.87 \\
& fixed-10k & 28.61 & 28.80 ± 9.87 & 91.86 ± 3.09 & 8269.06 & 1554.73 \\
& fixed-25k & 32.63 & 68.13 ± 22.77 & 79.60 ± 7.10 & 20475.06 & 2249.00 \\
& fixed-50k & 30.89 & 86.88 ± 20.14 & 58.46 ± 8.58 & 40152.73 & 3203.13 \\
& full-context & 29.53 & 100.00 ± 0.00 & 0.00 ± 0.00 & 86751.63 & 3767.70 \\
 \cmidrule{2-7}& \textsc{self-route} & \textbf{27.01} & 34.19 ± 35.59 & 74.68 ± 40.17 & 22726.84 & 1090.31 \\
& adaptive-\textit{k} & 25.68 & 66.16 ± 36.90 & 73.86 ± 23.04 & 25778.38 & 2818.23 \\
 \midrule\multirow{9}{4em}{info50k} & zeroshot & 5.28 & 0.00 ± 0.00 & 100.00 ± 0.00 & 58.13 & 43.59 \\
& fixed-1k & 11.73 & 1.47 ± 0.55 & 99.22 ± 0.23 & 1006.91 & 310.46 \\
& fixed-5k & 20.88 & 7.54 ± 2.52 & 96.02 ± 1.12 & 4235.36 & 511.71 \\
& fixed-10k & 21.53 & 15.39 ± 4.25 & 92.01 ± 2.13 & 8288.68 & 1556.82 \\
& fixed-25k & 25.29 & 39.55 ± 8.89 & 79.75 ± 5.38 & 20622.82 & 2609.31 \\
& fixed-50k & 30.18 & 76.90 ± 16.82 & 58.85 ± 9.99 & 41264.63 & 2885.44 \\
& full-context & 27.98 & 100.00 ± 0.00 & 0.00 ± 0.00 & 87936.41 & 3051.19 \\
 \cmidrule{2-7}& \textsc{self-route} & 19.57 & 26.38 ± 37.06 & 76.86 ± 38.66 & 21444.74 & 945.72 \\
& adaptive-\textit{k} & \textbf{22.93} & 67.43 ± 38.13 & 60.73 ± 25.94 & 39654.20 & 2781.81 \\
 \midrule
\multirow{9}{4em}{Average} & zeroshot & 6.43 & 0.00 & 0.00 & 58.13 & 43.84 \\
& fixed-1k & 19.78 & 5.70 & 99.20 & 1002.61 & 322.56 \\
& fixed-5k & 29.86 & 26.32 & 95.92 & 4218.45 & 773.33 \\
& fixed-10k & 32.12 & 42.49 & 91.85 & 8201.10 & 1406.36 \\
& fixed-25k & 33.54 & 66.09 & 79.62 & 20127.94 & 2016.19 \\
& fixed-50k & 32.18 & 84.48 & 58.24 & 40019.90 & 2835.45 \\
& full-context & 30.47 & 100.00 & 0.00 & 86677.57 & 3584.26 \\ \cmidrule{2-7}
& \textsc{self-route} & 26.61 & 44.55 & 76.00 & 21711.12 & 1447.15 \\
& adaptive-\textit{k} & \textbf{30.66} & 69.47 & 71.97 & 27572.90 & 2539.93 \\
\bottomrule
\end{tabular}
\caption{Full GPT-4o-mini's results in the HoloBench tasks.}
\label{tab:holobench-gpt-4o-mini}
\end{table*}

%% file: results/holobench-gpt-4o.tex
\begin{table*}[h]
\centering
\setlength{\tabcolsep}{6pt}
\small
\begin{tabular}{
l  
l  
r  
r  
r  
r  
r  
}
\toprule
Info amount & Method & Score & Context recall & Reduction (\%) & $n_\text{in}$ & $n_\text{out}$ \\
\midrule
\multirow{9}{4em}{info5k} & zeroshot & 7.22 & 0.00 ± 0.00 & 100.00 ± 0.00 & 58.13 & 65.21 \\
& fixed-1k & 26.14 & 12.05 ± 6.42 & 99.18 ± 0.43 & 1000.07 & 289.39 \\
& fixed-5k & 42.32 & 51.92 ± 29.82 & 95.85 ± 1.86 & 4194.01 & 913.32 \\
& fixed-10k & 49.79 & 66.68 ± 30.73 & 91.80 ± 2.92 & 8011.64 & 1334.67 \\
& fixed-25k & 46.27 & 78.48 ± 26.96 & 79.59 ± 4.56 & 19574.94 & 2087.61 \\
& fixed-50k & 43.82 & 86.79 ± 20.88 & 57.76 ± 5.92 & 39224.80 & 3188.69 \\
& full-context & 48.30 & 100.00 ± 0.00 & 0.00 ± 0.00 & 73652.50 & 3680.13 \\
 \cmidrule{2-7}& \textsc{self-route} & 41.86 & 65.79 ± 27.76 & 76.70 ± 38.60 & 17260.98 & 1139.29 \\
& adaptive-\textit{k} & \textbf{48.60} & 75.74 ± 30.48 & 74.07 ± 25.68 & 24625.02 & 1362.71 \\
 \midrule\multirow{9}{4em}{info10k} & zeroshot & 5.11 & 0.00 ± 0.00 & 100.00 ± 0.00 & 58.13 & 64.00 \\
& fixed-1k & 21.83 & 6.53 ± 3.11 & 99.19 ± 0.40 & 1003.83 & 298.53 \\
& fixed-5k & 32.45 & 31.77 ± 14.86 & 95.84 ± 1.90 & 4228.71 & 1001.83 \\
& fixed-10k & 36.48 & 59.10 ± 27.14 & 91.74 ± 3.47 & 8235.02 & 2589.58 \\
& fixed-25k & 39.65 & 78.18 ± 26.55 & 79.55 ± 5.42 & 19838.92 & 3527.68 \\
& fixed-50k & 38.55 & 87.34 ± 20.51 & 57.88 ± 6.41 & 39437.42 & 4061.82 \\
& full-context & 41.75 & 100.00 ± 0.00 & 0.00 ± 0.00 & 75768.00 & 4767.44 \\
 \cmidrule{2-7}& \textsc{self-route} & 32.37 & 45.04 ± 25.18 & 79.96 ± 36.00 & 17208.18 & 1146.12 \\
& adaptive-\textit{k} & \textbf{37.06} & 68.54 ± 32.55 & 79.22 ± 21.59 & 20233.99 & 2252.20 \\
 \midrule\multirow{9}{4em}{info25k} & zeroshot & 3.54 & 0.00 ± 0.00 & 100.00 ± 0.00 & 58.13 & 65.69 \\
& fixed-1k & 16.51 & 2.77 ± 1.15 & 99.21 ± 0.32 & 999.62 & 282.18 \\
& fixed-5k & 27.52 & 14.06 ± 5.50 & 95.96 ± 1.56 & 4215.72 & 1350.80 \\
& fixed-10k & 29.10 & 28.80 ± 9.87 & 91.86 ± 3.09 & 8269.06 & 2520.64 \\
& fixed-25k & 40.25 & 68.13 ± 22.77 & 79.60 ± 7.10 & 20475.06 & 3406.40 \\
& fixed-50k & 34.18 & 86.88 ± 20.14 & 58.46 ± 8.58 & 40152.73 & 4366.80 \\
& full-context & 42.24 & 100.00 ± 0.00 & 0.00 ± 0.00 & 75787.37 & 4802.50 \\
 \cmidrule{2-7}& \textsc{self-route} & 25.71 & 30.42 ± 32.70 & 77.87 ± 37.82 & 18525.29 & 1759.81 \\
& adaptive-\textit{k} & \textbf{33.46} & 66.16 ± 36.90 & 73.86 ± 23.04 & 25778.38 & 4017.11 \\
 \midrule\multirow{9}{4em}{info50k} & zeroshot & 5.19 & 0.00 ± 0.00 & 100.00 ± 0.00 & 58.13 & 62.30 \\
& fixed-1k & 11.59 & 1.47 ± 0.55 & 99.22 ± 0.23 & 1006.91 & 274.89 \\
& fixed-5k & 20.62 & 7.54 ± 2.52 & 96.02 ± 1.12 & 4235.36 & 735.80 \\
& fixed-10k & 23.15 & 15.39 ± 4.25 & 92.01 ± 2.13 & 8288.68 & 1595.42 \\
& fixed-25k & 28.11 & 39.55 ± 8.89 & 79.75 ± 5.38 & 20622.82 & 4269.89 \\
& fixed-50k & 34.58 & 76.90 ± 16.82 & 58.85 ± 9.99 & 41264.63 & 4244.67 \\
& full-context & 27.40 & 100.00 ± 0.00 & 0.00 ± 0.00 & 87936.41 & 3051.19 \\
 \cmidrule{2-7}& \textsc{self-route} & 19.28 & 21.54 ± 31.97 & 80.09 ± 36.03 & 15426.00 & 917.82 \\
& adaptive-\textit{k} & \textbf{30.10} & 67.43 ± 38.13 & 60.73 ± 25.94 & 39654.20 & 4374.22 \\
 \midrule
\multirow{9}{4em}{Average} & zeroshot & 5.26 & 0.00 & 0.00 & 58.13 & 64.30 \\
& fixed-1k & 19.02 & 5.70 & 99.20 & 1002.61 & 286.25 \\
& fixed-5k & 30.73 & 26.32 & 95.92 & 4218.45 & 1000.44 \\
& fixed-10k & 34.63 & 42.49 & 91.85 & 8201.10 & 2010.08 \\
& fixed-25k & 38.57 & 66.09 & 79.62 & 20127.94 & 3322.89 \\
& fixed-50k & 37.78 & 84.48 & 58.24 & 40019.90 & 3965.49 \\
& full-context & 39.92 & 100.00 & 0.00 & 78286.07 & 4075.32 \\ \cmidrule{2-7}
& \textsc{self-route} & 29.80 & 40.70 & 78.65 & 17105.11 & 1240.76 \\
& adaptive-\textit{k} & \textbf{37.30} & 69.47 & 71.97 & 27572.90 & 3001.56 \\
\bottomrule
\end{tabular}
\caption{Full GPT-4o's results in the HoloBench tasks.}
\label{tab:holobench-gpt-4o}
\end{table*}

%% file: results/holobench-gemini.tex
\begin{table*}[h]
\centering
\setlength{\tabcolsep}{6pt}
\small
\begin{tabular}{
l  
l  
r  
r  
r  
r  
r  
}
\toprule
Info amount & Method & Score & Context recall & Reduction (\%) & $n_\text{in}$ & $n_\text{out}$ \\
\midrule
\multirow{9}{4em}{info5k} & zeroshot & 10.19 & 0.00 ± 0.00 & 100.00 ± 0.00 & 52.64 & 644.08 \\
& fixed-1k & 27.78 & 12.05 ± 6.42 & 99.18 ± 0.43 & 1091.11 & 478.93 \\
& fixed-5k & 49.11 & 51.92 ± 29.82 & 95.85 ± 1.86 & 4610.40 & 1470.04 \\
& fixed-10k & 54.52 & 66.68 ± 30.73 & 91.80 ± 2.92 & 8760.92 & 2743.83 \\
& fixed-25k & 55.31 & 78.48 ± 26.96 & 79.59 ± 4.56 & 21300.94 & 3422.06 \\
& fixed-50k & 56.37 & 86.79 ± 20.88 & 57.76 ± 5.92 & 42764.79 & 3995.68 \\
& full-context & 63.27 & 100.00 ± 0.00 & 0.00 ± 0.00 & 94227.37 & 4584.58 \\
 \cmidrule{2-7}& \textsc{self-route} & 47.56 & 57.27 ± 31.23 & 88.38 ± 25.87 & 11559.12 & 1894.97 \\
& adaptive-\textit{k} & \textbf{55.68} & 75.74 ± 30.48 & 74.07 ± 25.68 & 26762.53 & 2776.24 \\
 \midrule\multirow{9}{4em}{info10k} & zeroshot & 8.33 & 0.00 ± 0.00 & 100.00 ± 0.00 & 52.64 & 572.52 \\
& fixed-1k & 21.09 & 6.53 ± 3.11 & 99.19 ± 0.40 & 1099.17 & 469.12 \\
& fixed-5k & 34.94 & 31.77 ± 14.86 & 95.84 ± 1.90 & 4683.90 & 2107.79 \\
& fixed-10k & 50.51 & 59.10 ± 27.14 & 91.74 ± 3.47 & 9105.06 & 2855.68 \\
& fixed-25k & 55.24 & 78.18 ± 26.55 & 79.55 ± 5.42 & 21717.67 & 4696.43 \\
& fixed-50k & 54.06 & 87.34 ± 20.51 & 57.88 ± 6.41 & 43101.23 & 6040.41 \\
& full-context & 53.65 & 100.00 ± 0.00 & 0.00 ± 0.00 & 94626.73 & 6381.93 \\
 \cmidrule{2-7}& \textsc{self-route} & 35.72 & 36.72 ± 22.19 & 89.40 ± 24.10 & 10196.77 & 2031.94 \\
& adaptive-\textit{k} & \textbf{56.26} & 68.54 ± 32.55 & 79.22 ± 21.59 & 22081.99 & 3637.80 \\
 \midrule\multirow{9}{4em}{info25k} & zeroshot & 6.28 & 0.00 ± 0.00 & 100.00 ± 0.00 & 52.64 & 657.34 \\
& fixed-1k & 15.42 & 2.77 ± 1.15 & 99.21 ± 0.32 & 1098.33 & 455.64 \\
& fixed-5k & 31.18 & 14.06 ± 5.50 & 95.96 ± 1.56 & 4699.37 & 1695.57 \\
& fixed-10k & 37.86 & 28.80 ± 9.87 & 91.86 ± 3.09 & 9230.02 & 3053.66 \\
& fixed-25k & 42.87 & 68.13 ± 22.77 & 79.60 ± 7.10 & 22790.66 & 6461.89 \\
& fixed-50k & 44.54 & 86.88 ± 20.14 & 58.46 ± 8.58 & 44317.83 & 7982.89 \\
& full-context & 42.19 & 100.00 ± 0.00 & 0.00 ± 0.00 & 95716.53 & 9289.30 \\
 \cmidrule{2-7}& \textsc{self-route} & 28.12 & 20.06 ± 22.11 & 89.55 ± 24.11 & 11120.20 & 2086.64 \\
& adaptive-\textit{k} & \textbf{43.76} & 66.16 ± 36.90 & 73.86 ± 23.04 & 28207.51 & 5901.29 \\
 \midrule\multirow{9}{4em}{info50k} & zeroshot & 6.95 & 0.00 ± 0.00 & 100.00 ± 0.00 & 52.64 & 735.01 \\
& fixed-1k & 9.77 & 1.47 ± 0.55 & 99.22 ± 0.23 & 1108.12 & 445.62 \\
& fixed-5k & 22.66 & 7.54 ± 2.52 & 96.02 ± 1.12 & 4730.58 & 1836.44 \\
& fixed-10k & 30.00 & 15.39 ± 4.25 & 92.01 ± 2.13 & 9277.87 & 2990.94 \\
& fixed-25k & 33.71 & 39.55 ± 8.89 & 79.75 ± 5.38 & 23092.23 & 6596.67 \\
& fixed-50k & 35.68 & 76.90 ± 16.82 & 58.85 ± 9.99 & 46101.70 & 8373.53 \\
& full-context & 45.44 & 100.00 ± 0.00 & 0.00 ± 0.00 & 97597.56 & 8792.94 \\
 \cmidrule{2-7}& \textsc{self-route} & 22.37 & 11.75 ± 19.29 & 91.76 ± 19.93 & 8923.41 & 1610.90 \\
& adaptive-\textit{k} & \textbf{31.72} & 67.43 ± 38.13 & 60.73 ± 25.94 & 43888.98 & 7643.86 \\
 \midrule
\multirow{9}{4em}{Average} & zeroshot & 7.94 & 0.00 & 0.00 & 52.64 & 652.24 \\
& fixed-1k & 18.51 & 5.70 & 99.20 & 1099.18 & 462.33 \\
& fixed-5k & 34.47 & 26.32 & 95.92 & 4681.06 & 1777.46 \\
& fixed-10k & 43.22 & 42.49 & 91.85 & 9093.47 & 2911.03 \\
& fixed-25k & 46.78 & 66.09 & 79.62 & 22225.38 & 5294.26 \\
& fixed-50k & 47.66 & 84.48 & 58.24 & 44071.39 & 6598.13 \\
& full-context & 51.14 & 100.00 & 0.00 & 95542.05 & 7262.19 \\ \cmidrule{2-7}
& \textsc{self-route} & 33.44 & 31.45 & 89.78 & 10449.88 & 1906.11 \\
& adaptive-\textit{k} & \textbf{46.85} & 69.47 & 71.97 & 30235.25 & 4989.80 \\
\bottomrule
\end{tabular}
\caption{Full Gemini-2.5-Flash's results in the HoloBench tasks.}
\label{tab:holobench-gemini}
\end{table*}

%% file: results/holobench-llama4-scout.tex
\begin{table*}[h]
\centering
\setlength{\tabcolsep}{6pt}
\small
\begin{tabular}{
l  
l  
r  
r  
r  
r  
r  
}
\toprule
Info amount & Method & Score & Context recall & Reduction (\%) & $n_\text{in}$ & $n_\text{out}$ \\
\midrule
\multirow{9}{4em}{info5k} & zeroshot & 9.49 & 0.00 ± 0.00 & 100.00 ± 0.00 & 56.18 & 266.46 \\
& fixed-1k & 29.52 & 12.05 ± 6.42 & 99.18 ± 0.43 & 994.79 & 450.14 \\
& fixed-5k & 40.38 & 51.92 ± 29.82 & 95.85 ± 1.86 & 4195.84 & 650.72 \\
& fixed-10k & 40.73 & 66.68 ± 30.73 & 91.80 ± 2.92 & 8024.09 & 687.39 \\
& fixed-25k & 37.33 & 78.48 ± 26.96 & 79.59 ± 4.56 & 19625.13 & 711.01 \\
& fixed-50k & 34.47 & 86.79 ± 20.88 & 57.76 ± 5.92 & 39342.62 & 707.11 \\
& full-context & 36.32 & 100.00 ± 0.00 & 0.00 ± 0.00 & 86093.44 & 789.47 \\
 \cmidrule{2-7}& \textsc{self-route} & 36.35 & 70.18 ± 29.18 & 69.25 ± 43.22 & 27700.34 & 785.69 \\
& adaptive-\textit{k} & \textbf{39.01} & 75.74 ± 30.48 & 74.07 ± 25.68 & 24711.58 & 1170.32 \\
 \midrule\multirow{9}{4em}{info10k} & zeroshot & 7.01 & 0.00 ± 0.00 & 100.00 ± 0.00 & 56.18 & 260.24 \\
& fixed-1k & 19.25 & 6.53 ± 3.11 & 99.19 ± 0.40 & 999.90 & 473.22 \\
& fixed-5k & 33.58 & 31.77 ± 14.86 & 95.84 ± 1.90 & 4233.97 & 708.81 \\
& fixed-10k & 31.87 & 59.10 ± 27.14 & 91.74 ± 3.47 & 8254.73 & 796.22 \\
& fixed-25k & 29.75 & 78.18 ± 26.55 & 79.55 ± 5.42 & 19898.58 & 840.93 \\
& fixed-50k & 28.60 & 87.34 ± 20.51 & 57.88 ± 6.41 & 39561.99 & 1070.07 \\
& full-context & 30.50 & 100.00 ± 0.00 & 0.00 ± 0.00 & 86347.34 & 1143.99 \\
 \cmidrule{2-7}& \textsc{self-route} & 33.51 & 51.87 ± 29.27 & 72.57 ± 41.54 & 25307.50 & 1167.84 \\
& adaptive-\textit{k} & \textbf{34.69} & 68.54 ± 32.55 & 79.22 ± 21.59 & 20348.23 & 709.23 \\
 \midrule\multirow{9}{4em}{info25k} & zeroshot & 7.23 & 0.00 ± 0.00 & 100.00 ± 0.00 & 56.18 & 277.41 \\
& fixed-1k & 17.62 & 2.77 ± 1.15 & 99.21 ± 0.32 & 997.53 & 473.57 \\
& fixed-5k & 29.44 & 14.06 ± 5.50 & 95.96 ± 1.56 & 4226.03 & 682.94 \\
& fixed-10k & 28.95 & 28.80 ± 9.87 & 91.86 ± 3.09 & 8298.52 & 759.61 \\
& fixed-25k & 31.39 & 68.13 ± 22.77 & 79.60 ± 7.10 & 20562.19 & 795.73 \\
& fixed-50k & 28.35 & 86.88 ± 20.14 & 58.46 ± 8.58 & 40309.13 & 1268.03 \\
& full-context & 25.94 & 100.00 ± 0.00 & 0.00 ± 0.00 & 86997.00 & 961.38 \\
 \cmidrule{2-7}& \textsc{self-route} & \textbf{29.08} & 32.25 ± 34.30 & 76.81 ± 38.65 & 21041.81 & 914.91 \\
& adaptive-\textit{k} & 26.90 & 66.16 ± 36.90 & 73.86 ± 23.04 & 25958.96 & 854.74 \\
 \midrule\multirow{9}{4em}{info50k} & zeroshot & 6.18 & 0.00 ± 0.00 & 100.00 ± 0.00 & 56.18 & 427.23 \\
& fixed-1k & 11.93 & 1.47 ± 0.55 & 99.22 ± 0.23 & 1004.93 & 416.12 \\
& fixed-5k & 22.89 & 7.54 ± 2.52 & 96.02 ± 1.12 & 4246.40 & 639.41 \\
& fixed-10k & 23.75 & 15.39 ± 4.25 & 92.01 ± 2.13 & 8317.29 & 735.40 \\
& fixed-25k & 23.94 & 39.55 ± 8.89 & 79.75 ± 5.38 & 20706.06 & 956.62 \\
& fixed-50k & 25.46 & 76.90 ± 16.82 & 58.85 ± 9.99 & 41427.09 & 854.82 \\
& full-context & 22.10 & 100.00 ± 0.00 & 0.00 ± 0.00 & 88197.90 & 1321.16 \\
 \cmidrule{2-7}& \textsc{self-route} & 22.24 & 32.58 ± 40.92 & 70.50 ± 42.76 & 27508.79 & 980.10 \\
& adaptive-\textit{k} & \textbf{25.45} & 67.43 ± 38.13 & 60.73 ± 25.94 & 39897.64 & 1182.12 \\
 \midrule
\multirow{9}{4em}{Average} & zeroshot & 7.48 & 0.00 & 0.00 & 56.18 & 307.84 \\
& fixed-1k & 19.58 & 5.70 & 99.20 & 999.29 & 453.26 \\
& fixed-5k & 31.57 & 26.32 & 95.92 & 4225.56 & 670.47 \\
& fixed-10k & 31.33 & 42.49 & 91.85 & 8223.66 & 744.66 \\
& fixed-25k & 30.60 & 66.09 & 79.62 & 20197.99 & 826.07 \\
& fixed-50k & 29.22 & 84.48 & 58.24 & 40160.21 & 975.01 \\
& full-context & 28.72 & 100.00 & 0.00 & 86908.92 & 1054.00 \\ \cmidrule{2-7}
& \textsc{self-route} & 30.29 & 46.72 & 72.28 & 25389.61 & 962.14 \\
& adaptive-\textit{k} & \textbf{31.51} & 69.47 & 71.97 & 27729.10 & 979.11 \\
\bottomrule
\end{tabular}
\caption{Full Llama4-Scout's results in the HoloBench tasks.}
\label{tab:holobench-llama4-scout}
\end{table*}

%% file: results/holobench-llama4-maverick.tex
\begin{table*}[h]
\centering
\setlength{\tabcolsep}{6pt}
\small
\begin{tabular}{
l  
l  
r  
r  
r  
r  
r  
}
\toprule
Info amount & Method & Score & Context recall & Reduction (\%) & $n_\text{in}$ & $n_\text{out}$ \\
\midrule
\multirow{9}{4em}{info5k} & zeroshot & 9.65 & 0.00 ± 0.00 & 100.00 ± 0.00 & 56.18 & 328.77 \\
& fixed-1k & 28.05 & 12.05 ± 6.42 & 99.18 ± 0.43 & 994.79 & 591.14 \\
& fixed-5k & 48.57 & 51.92 ± 29.82 & 95.85 ± 1.86 & 4195.84 & 799.80 \\
& fixed-10k & 54.30 & 66.68 ± 30.73 & 91.80 ± 2.92 & 8024.09 & 834.79 \\
& fixed-25k & 56.39 & 78.48 ± 26.96 & 79.59 ± 4.56 & 19625.13 & 874.41 \\
& fixed-50k & 53.54 & 86.79 ± 20.88 & 57.76 ± 5.92 & 39342.62 & 919.02 \\
& full-context & 55.13 & 100.00 ± 0.00 & 0.00 ± 0.00 & 86093.44 & 1101.66 \\
 \cmidrule{2-7}& \textsc{self-route} & 48.47 & 65.03 ± 29.24 & 79.91 ± 35.98 & 18195.21 & 897.80 \\
& adaptive-\textit{k} & \textbf{51.77} & 75.74 ± 30.48 & 74.07 ± 25.68 & 24711.58 & 836.68 \\
 \midrule\multirow{9}{4em}{info10k} & zeroshot & 9.06 & 0.00 ± 0.00 & 100.00 ± 0.00 & 56.18 & 322.59 \\
& fixed-1k & 23.16 & 6.53 ± 3.11 & 99.19 ± 0.40 & 999.90 & 609.29 \\
& fixed-5k & 39.02 & 31.77 ± 14.86 & 95.84 ± 1.90 & 4233.97 & 892.93 \\
& fixed-10k & 42.91 & 59.10 ± 27.14 & 91.74 ± 3.47 & 8254.73 & 927.12 \\
& fixed-25k & 44.54 & 78.18 ± 26.55 & 79.55 ± 5.42 & 19898.58 & 883.56 \\
& fixed-50k & 50.37 & 87.34 ± 20.51 & 57.88 ± 6.41 & 39561.99 & 1020.78 \\
& full-context & 48.02 & 100.00 ± 0.00 & 0.00 ± 0.00 & 86347.34 & 1357.50 \\
 \cmidrule{2-7}& \textsc{self-route} & 37.68 & 44.92 ± 27.75 & 79.95 ± 36.00 & 18010.20 & 1053.86 \\
& adaptive-\textit{k} & \textbf{44.91} & 68.54 ± 32.55 & 79.22 ± 21.59 & 20348.23 & 934.27 \\
 \midrule\multirow{9}{4em}{info25k} & zeroshot & 7.26 & 0.00 ± 0.00 & 100.00 ± 0.00 & 56.18 & 322.77 \\
& fixed-1k & 19.15 & 2.77 ± 1.15 & 99.21 ± 0.32 & 997.53 & 629.03 \\
& fixed-5k & 30.36 & 14.06 ± 5.50 & 95.96 ± 1.56 & 4226.03 & 896.96 \\
& fixed-10k & 30.28 & 28.80 ± 9.87 & 91.86 ± 3.09 & 8298.52 & 867.99 \\
& fixed-25k & 40.47 & 68.13 ± 22.77 & 79.60 ± 7.10 & 20562.19 & 1053.26 \\
& fixed-50k & 44.69 & 86.88 ± 20.14 & 58.46 ± 8.58 & 40309.13 & 1291.90 \\
& full-context & 43.38 & 100.00 ± 0.00 & 0.00 ± 0.00 & 86997.00 & 1455.21 \\
 \cmidrule{2-7}& \textsc{self-route} & 28.57 & 29.97 ± 33.04 & 79.00 ± 36.97 & 19123.37 & 1038.17 \\
& adaptive-\textit{k} & \textbf{39.40} & 66.16 ± 36.90 & 73.86 ± 23.04 & 25958.96 & 1122.81 \\
 \midrule\multirow{9}{4em}{info50k} & zeroshot & 8.51 & 0.00 ± 0.00 & 100.00 ± 0.00 & 56.18 & 342.58 \\
& fixed-1k & 11.24 & 1.47 ± 0.55 & 99.22 ± 0.23 & 1004.93 & 574.83 \\
& fixed-5k & 22.39 & 7.54 ± 2.52 & 96.02 ± 1.12 & 4246.40 & 783.99 \\
& fixed-10k & 24.53 & 15.39 ± 4.25 & 92.01 ± 2.13 & 8317.29 & 825.16 \\
& fixed-25k & 27.94 & 39.55 ± 8.89 & 79.75 ± 5.38 & 20706.06 & 1127.69 \\
& fixed-50k & 34.89 & 76.90 ± 16.82 & 58.85 ± 9.99 & 41427.09 & 1317.22 \\
& full-context & 36.54 & 100.00 ± 0.00 & 0.00 ± 0.00 & 88197.90 & 1706.96 \\
 \cmidrule{2-7}& \textsc{self-route} & 21.90 & 24.17 ± 35.52 & 78.99 ± 36.95 & 19460.93 & 1183.29 \\
& adaptive-\textit{k} & \textbf{34.63} & 67.43 ± 38.13 & 60.73 ± 25.94 & 39897.64 & 1687.78 \\
 \midrule
\multirow{9}{4em}{Average} & zeroshot & 8.62 & 0.00 & 0.00 & 56.18 & 329.18 \\
& fixed-1k & 20.40 & 5.70 & 99.20 & 999.29 & 601.07 \\
& fixed-5k & 35.09 & 26.32 & 95.92 & 4225.56 & 843.42 \\
& fixed-10k & 38.00 & 42.49 & 91.85 & 8223.66 & 863.76 \\
& fixed-25k & 42.33 & 66.09 & 79.62 & 20197.99 & 984.73 \\
& fixed-50k & 45.87 & 84.48 & 58.24 & 40160.21 & 1137.23 \\
& full-context & 45.77 & 100.00 & 0.00 & 86908.92 & 1405.33 \\ \cmidrule{2-7}
& \textsc{self-route} & 34.16 & 41.02 & 79.46 & 18697.43 & 1043.28 \\
& adaptive-\textit{k} & \textbf{42.68} & 69.47 & 71.97 & 27729.10 & 1145.38 \\
\bottomrule
\end{tabular}
\caption{Full Llama4-Maverick's results in the HoloBench tasks.}
\label{tab:holobench-llama4-maverick}
\end{table*}

%% file: results/qa-gpt-4o-gte.tex
\begin{table*}[h]
\centering
\setlength{\tabcolsep}{6pt}
\small
\begin{tabular}{
l  
l  
r  
r  
r  
r  
r  
}
\toprule
Task & Method & Score & Context recall & Reduction (\%) & $n_\text{in}$ & $n_\text{out}$ \\
\midrule
\multirow{9}{4em}{HotpotQA} & zeroshot & 50 & 0.00 ± 0.00 & 100.00 ± 0.00 & 63.03 ± 6.98 & 19.31 ± 16.53 \\
& fixed-1k & 50 & 6.00 ± 16.33 & 99.16 ± 0.20 & 881.29 ± 76.42 & 31.19 ± 16.09 \\
& fixed-5k & 51 & 9.33 ± 19.44 & 95.84 ± 0.67 & 4106.52 ± 327.20 & 30.61 ± 16.31 \\
& fixed-10k & 54 & 13.67 ± 23.73 & 91.81 ± 1.19 & 8145.75 ± 576.29 & 29.29 ± 15.20 \\
& fixed-25k & 58 & 27.50 ± 32.17 & 80.21 ± 2.40 & 20337.90 ± 1369.28 & 29.06 ± 17.08 \\
& fixed-50k & 60 & 43.50 ± 33.16 & 61.79 ± 3.81 & 40520.53 ± 2661.25 & 28.96 ± 18.42 \\
& full-context & 48 & 100.00 ± 0.00 & 0.00 ± 0.00 & 109666.92 ± 5537.59 & 18.91 ± 20.30 \\
 \cmidrule{2-7}& \textsc{self-route} & 46 & 25.83 ± 39.17 & 78.64 ± 37.04 & 76201.89 ± 52199.49 & 22.69 ± 20.51 \\
& adaptive-\textit{k} & \textbf{49} & 5.50 ± 15.72 & 99.20 ± 0.36 & 877.06 ± 409.16 & 32.87 ± 23.82 \\
 \midrule\multirow{9}{4em}{NQ} & zeroshot & 57 & 0.00 ± 0.00 & 100.00 ± 0.00 & 53.37 ± 2.29 & 27.38 ± 24.23 \\
& fixed-1k & 53 & 2.65 ± 9.56 & 99.29 ± 0.19 & 813.57 ± 84.46 & 31.61 ± 25.74 \\
& fixed-5k & 53 & 9.07 ± 18.72 & 96.44 ± 0.49 & 3871.35 ± 353.77 & 33.67 ± 28.72 \\
& fixed-10k & 58 & 14.22 ± 22.67 & 92.96 ± 0.77 & 7702.16 ± 668.50 & 34.24 ± 26.91 \\
& fixed-25k & 62 & 31.38 ± 31.59 & 82.54 ± 1.61 & 19183.81 ± 1661.81 & 35.55 ± 29.97 \\
& fixed-50k & 64 & 45.18 ± 33.28 & 65.32 ± 2.88 & 38288.26 ± 3141.73 & 36.84 ± 34.84 \\
& full-context & 41 & 100.00 ± 0.00 & 0.00 ± 0.00 & 110607.54 ± 4711.16 & 24.74 ± 33.79 \\
 \cmidrule{2-7}& \textsc{self-route} & 49 & 20.67 ± 31.77 & 78.12 ± 38.03 & 54951.43 ± 55725.18 & 27.13 ± 25.83 \\
& adaptive-\textit{k} & \textbf{51} & 2.85 ± 9.72 & 99.22 ± 0.24 & 907.79 ± 263.30 & 32.75 ± 25.98 \\
 \midrule\multirow{9}{4em}{TriviaQA} & zeroshot & 91 & 0.00 ± 0.00 & 100.00 ± 0.00 & 60.09 ± 7.79 & 8.33 ± 7.16 \\
& fixed-1k & 96 & 3.00 ± 13.48 & 99.15 ± 0.49 & 857.89 ± 96.49 & 16.98 ± 10.66 \\
& fixed-5k & 94 & 4.79 ± 15.73 & 95.94 ± 0.83 & 4040.80 ± 358.14 & 17.35 ± 10.37 \\
& fixed-10k & 93 & 6.43 ± 17.79 & 92.04 ± 1.25 & 8038.26 ± 670.72 & 17.64 ± 10.47 \\
& fixed-25k & 96 & 16.65 ± 30.03 & 80.93 ± 1.99 & 19985.92 ± 1507.53 & 16.64 ± 10.50 \\
& fixed-50k & 95 & 39.57 ± 41.22 & 63.01 ± 3.13 & 39806.06 ± 2867.46 & 17.17 ± 10.11 \\
& full-context & 62 & 100.00 ± 0.00 & 0.00 ± 0.00 & 110733.69 ± 3419.97 & 12.94 ± 12.48 \\
 \cmidrule{2-7}& \textsc{self-route} & 88 & 10.20 ± 24.97 & 87.30 ± 27.60 & 32620.05 ± 48431.50 & 13.76 ± 10.53 \\
& adaptive-\textit{k} & \textbf{93} & 3.00 ± 13.48 & 99.16 ± 0.56 & 929.99 ± 604.34 & 17.33 ± 10.55 \\
 \midrule
\multirow{9}{4em}{Average} & zeroshot & 66.00 & 0.00 & 0.00 & 58.83 & 18.34 \\
& fixed-1k & 66.33 & 3.88 & 99.20 & 850.92 & 26.59 \\
& fixed-5k & 66.00 & 7.73 & 96.07 & 4006.22 & 27.21 \\
& fixed-10k & 68.33 & 11.44 & 92.27 & 7962.06 & 27.06 \\
& fixed-25k & 72.00 & 25.18 & 81.23 & 19835.88 & 27.08 \\
& fixed-50k & 73.00 & 42.75 & 63.38 & 39538.28 & 27.66 \\
& full-context & 50.33 & 100.00 & 0.00 & 110336.05 & 18.86 \\ \cmidrule{2-7}
& \textsc{self-route} & 61.00 & 18.90 & 81.35 & 54591.12 & 21.19 \\
& adaptive-\textit{k} & \textbf{64.33} & 3.78 & 99.19 & 904.95 & 27.65 \\
\bottomrule
\end{tabular}
\caption{Full GPT-4o's results in the factoid QA tasks with the embeddings by gte-Qwen2-1.5B-instruct.}
\label{tab:qa-gpt-4o-gte}
\end{table*}

%% file: results/holobench-gpt-4o-gte.tex
\begin{table*}[h]
\centering
\setlength{\tabcolsep}{6pt}
\small
\begin{tabular}{
l  
l  
r  
r  
r  
r  
r  
}
\toprule
Info amount & Method & Score & Context recall & Reduction (\%) & $n_\text{in}$ & $n_\text{out}$ \\
\midrule
\multirow{9}{4em}{info5k} & zeroshot & 7.22 & 0.00 ± 0.00 & 100.00 ± 0.00 & 58.13 & 65.21 \\
& fixed-1k & 26.14 & 12.05 ± 6.42 & 99.18 ± 0.43 & 1000.07 & 289.39 \\
& fixed-5k & 42.32 & 51.92 ± 29.82 & 95.85 ± 1.86 & 4194.01 & 913.32 \\
& fixed-10k & 49.79 & 66.68 ± 30.73 & 91.80 ± 2.92 & 8011.64 & 1334.67 \\
& fixed-25k & 46.27 & 78.48 ± 26.96 & 79.59 ± 4.56 & 19574.94 & 2087.61 \\
& fixed-50k & 43.82 & 86.79 ± 20.88 & 57.76 ± 5.92 & 39224.80 & 3188.69 \\
& full-context & 48.30 & 100.00 ± 0.00 & 0.00 ± 0.00 & 73652.50 & 3680.13 \\
 \cmidrule{2-7}& \textsc{self-route} & 40.69 & 65.18 ± 28.08 & 76.14 ± 38.33 & 12113.12 & 1288.40 \\
& adaptive-\textit{k} & \textbf{45.23} & 82.20 ± 25.15 & 64.79 ± 32.49 & 29079.80 & 1879.49 \\
 \midrule\multirow{9}{4em}{info10k} & zeroshot & 5.11 & 0.00 ± 0.00 & 100.00 ± 0.00 & 58.13 & 64.00 \\
& fixed-1k & 21.83 & 6.53 ± 3.11 & 99.19 ± 0.40 & 1003.83 & 298.53 \\
& fixed-5k & 32.45 & 31.77 ± 14.86 & 95.84 ± 1.90 & 4228.71 & 1001.83 \\
& fixed-10k & 36.48 & 59.10 ± 27.14 & 91.74 ± 3.47 & 8235.02 & 2589.58 \\
& fixed-25k & 39.65 & 78.18 ± 26.55 & 79.55 ± 5.42 & 19838.92 & 3527.68 \\
& fixed-50k & 38.55 & 87.34 ± 20.51 & 57.88 ± 6.41 & 39437.42 & 4061.82 \\
& full-context & 41.75 & 100.00 ± 0.00 & 0.00 ± 0.00 & 75768.00 & 4767.44 \\
 \cmidrule{2-7}& \textsc{self-route} & 32.28 & 45.87 ± 24.77 & 79.47 ± 35.80 & 18646.59 & 1307.47 \\
& adaptive-\textit{k} & \textbf{39.66} & 78.99 ± 28.37 & 65.70 ± 30.96 & 28475.07 & 2238.23 \\
 \midrule\multirow{9}{4em}{info25k} & zeroshot & 3.54 & 0.00 ± 0.00 & 100.00 ± 0.00 & 58.13 & 65.69 \\
& fixed-1k & 16.51 & 2.77 ± 1.15 & 99.21 ± 0.32 & 999.62 & 282.18 \\
& fixed-5k & 27.52 & 14.06 ± 5.50 & 95.96 ± 1.56 & 4215.72 & 1350.80 \\
& fixed-10k & 29.10 & 28.80 ± 9.87 & 91.86 ± 3.09 & 8269.06 & 2520.64 \\
& fixed-25k & 40.25 & 68.13 ± 22.77 & 79.60 ± 7.10 & 20475.06 & 3406.40 \\
& fixed-50k & 34.18 & 86.88 ± 20.14 & 58.46 ± 8.58 & 40152.73 & 4366.80 \\
& full-context & 42.24 & 100.00 ± 0.00 & 0.00 ± 0.00 & 75787.37 & 4802.50 \\
 \cmidrule{2-7}& \textsc{self-route} & 23.60 & 31.02 ± 32.35 & 77.63 ± 37.70 & 18905.02 & 1334.77 \\
& adaptive-\textit{k} & \textbf{36.33} & 76.47 ± 31.62 & 58.43 ± 29.94 & 36711.21 & 3509.82 \\
 \midrule\multirow{9}{4em}{info50k} & zeroshot & 5.19 & 0.00 ± 0.00 & 100.00 ± 0.00 & 58.13 & 62.30 \\
& fixed-1k & 11.59 & 1.47 ± 0.55 & 99.22 ± 0.23 & 1006.91 & 274.89 \\
& fixed-5k & 20.62 & 7.54 ± 2.52 & 96.02 ± 1.12 & 4235.36 & 735.80 \\
& fixed-10k & 23.15 & 15.39 ± 4.25 & 92.01 ± 2.13 & 8288.68 & 1595.42 \\
& fixed-25k & 28.11 & 39.55 ± 8.89 & 79.75 ± 5.38 & 20622.82 & 4269.89 \\
& fixed-50k & 34.58 & 76.90 ± 16.82 & 58.85 ± 9.99 & 41264.63 & 4244.67 \\
& full-context & 27.40 & 100.00 ± 0.00 & 0.00 ± 0.00 & 87936.41 & 3051.19 \\
 \cmidrule{2-7}& \textsc{self-route} & 22.08 & 21.90 ± 31.81 & 79.88 ± 35.94 & 17563.86 & 1843.27 \\
& adaptive-\textit{k} & \textbf{30.80} & 72.54 ± 36.80 & 49.06 ± 29.83 & 46590.91 & 3806.21 \\
 \midrule
\multirow{9}{4em}{Average} & zeroshot & 5.26 & 0.00 & 0.00 & 58.13 & 64.30 \\
& fixed-1k & 19.02 & 5.70 & 99.20 & 1002.61 & 286.25 \\
& fixed-5k & 30.73 & 26.32 & 95.92 & 4218.45 & 1000.44 \\
& fixed-10k & 34.63 & 42.49 & 91.85 & 8201.10 & 2010.08 \\
& fixed-25k & 38.57 & 66.09 & 79.62 & 20127.94 & 3322.89 \\
& fixed-50k & 37.78 & 84.48 & 58.24 & 40019.90 & 3965.49 \\
& full-context & 39.92 & 100.00 & 0.00 & 78286.07 & 4075.32 \\ \cmidrule{2-7}
& \textsc{self-route} & 29.66 & 40.99 & 78.28 & 16807.15 & 1443.48 \\
& adaptive-\textit{k} & \textbf{38.00} & 77.55 & 59.49 & 35214.25 & 2858.44 \\
\bottomrule
\end{tabular}
\caption{Full GPT-4o's results in the HoloBench tasks with the embeddings by gte-Qwen2-1.5B-Instruct.}
\label{tab:holobench-gpt-4o-gte}
\end{table*}